%% file: alpa-serve.tex
\documentclass[letterpaper,twocolumn,10pt]{article}
\usepackage{usenix}
\usepackage[available,functional,reproduced]{usenixbadges}

\usepackage{tikz}
\usepackage{float}
\usepackage{amsmath}
\usepackage{balance}
\usepackage{booktabs}
\usepackage[capitalise]{cleveref}
\usepackage{xcolor}
\usepackage{xspace}
\usepackage{multirow}
\usepackage{tabularx}
\usepackage{algorithm}
\usepackage{algpseudocode}
\algtext*{EndWhile}
\algtext*{EndIf}
\algtext*{EndFor}
\usepackage{caption}
\usepackage{subcaption}
\usepackage[compact]{titlesec}

\usepackage{enumitem}
\setitemize{noitemsep,topsep=0pt,parsep=0pt,partopsep=0pt}

\newif\ifcomments
\commentstrue 
\ifcomments
    \newcommand{\hao}[1]{{\color{green}{\bf\sf [Hao: #1]}}}
    \newcommand{\ion}[1]{{\color{red}{\bf\sf [Ion: #1]}}}
    \newcommand{\joey}[1]{{\color{cyan}{\bf\sf [Joey: #1]}}}
    
    \newcommand{\lianmin}[1]{{\color{blue}{\bf\sf [Lianmin: #1]}}}
    \newcommand{\zhuohan}[1]{{\color{cyan}{\bf\sf [Zhuohan: #1]}}}
    \newcommand{\vincent}[1]{{\color{violet}{\bf\sf [Vincent: #1]}}}
    \newcommand{\yinmin}[1]{{\color{red}{\bf\sf [Yinmin: #1]}}}
    \newcommand{\ying}[1]{{\color{orange}{\bf\sf [Ying: #1]}}}
    \newcommand{\todo}[1]{{\color{orange}{\bf\sf [TODO: #1]}}}
    
\else
    \newcommand{\hao}[1]{}
    \newcommand{\ion}[1]{}
    \newcommand{\joey}[1]{}
    \newcommand{\lianmin}[1]{}
    \newcommand{\zhuohan}[1]{}
    \newcommand{\vincent}[1]{}
    \newcommand{\yinmin}[1]{}
    \newcommand{\ying}[1]{}
    \newcommand{\todo}[1]{}
    
\fi

\newcommand{\sys}[0]{AlpaServe\xspace}
\def\Snospace~{\S{}}

\newcommand{\heading}[1]{\vspace{4pt}\noindent\textbf{#1}}
\newcommand{\topheading}[1]{\noindent\textbf{#1}}

\begin{document}

\date{}

\title{\sys: Statistical Multiplexing with Model Parallelism \\ for Deep Learning Serving}

\author{
\rm{Zhuohan Li$^{\text{1}, *}$ \enskip
    Lianmin Zheng$^{\text{1}, *}$ \enskip
    Yinmin Zhong$^{\text{2}, *}$ \enskip
    Vincent Liu$^{\text{3}}$ \enskip
    Ying Sheng$^{\text{4}}$  \enskip}
\\
\rm{Xin Jin$^{\text{2}}$ \enskip
    Yanping Huang$^{\text{5}}$ \enskip
    Zhifeng Chen$^{\text{5}}$ \enskip
    Hao Zhang$^{\text{6}}$ \enskip
    Joseph E. Gonzalez$^{\text{1}}$ \enskip
    Ion Stoica$^{\text{1}}$ \enskip}\\
\\
{$^{\text{1}}$UC Berkeley\enskip $^{\text{2}}$Peking University\enskip $^{\text{3}}$University of Pennsylvania\enskip}
\\
{$^{\text{4}}$Stanford University\enskip $^{\text{5}}$Google\enskip $^{\text{6}}$UC San Diego\enskip}
}

\maketitle

{\let\thefootnote\relax\footnote{{$^*$Equal contribution.}}}
\input{abstract}
\input{sec.intro}
\input{sec.background}
\input{sec.motivation}
\input{sec.method}
\input{sec.eval}
\input{sec.related_work}
\input{sec.conclusion}

\balance
\bibliographystyle{plain}
\bibliography{alpa-serve}

\end{document}

%% file: abstract.tex
\begin{abstract}

Model parallelism is conventionally viewed as a method to scale a single large deep learning model beyond the memory limits of a single device. 
In this paper, we demonstrate that model parallelism can be additionally used for the statistical multiplexing of multiple devices when serving multiple models, even when a single model can fit into a single device. 
Our work reveals a fundamental trade-off between the overhead introduced by model parallelism and the opportunity to exploit statistical multiplexing to reduce serving latency in the presence of bursty workloads.
We explore the new trade-off space and present a novel serving system, \sys, that determines an efficient strategy for placing and parallelizing collections of large deep learning models across a distributed cluster.
Evaluation results on production workloads show that \sys can process requests at up to $10\times$ higher rates or $6\times$ more burstiness while staying within latency constraints for more than 99\% of requests.

\end{abstract}

%% file: sec.intro.tex
\section{Introduction}

Advances in self-supervised learning have enabled exponential scaling in model sizes. For example, large pretrained models like BERT~\cite{devlin2018bert} and GPT-3~\cite{brown2020language} have unlocked a plethora of new machine learning (ML) applications from Copilot~\cite{copilot} to copy.ai~\cite{copyai} and ChatGPT~\cite{chatgpt}.

Serving these very large models is challenging because of their high computational and memory requirements.
For example, 
GPT-3 requires 325\,GB of memory to store its parameters as well as a requisite amount of computation to run inference.
To serve this model, one would need at least 5 of Nvidia's newest Hopper 80\,GB GPUs just to hold the  weights and potentially many more to run in real-time.
Worse yet, the explosive growth of model sizes continues unabated~\cite{chowdhery2022palm,fedus2022switch}.
Techniques like model compression and pruning are not sufficient in face of the exponential growth in model sizes and often come at the expense of reduced model quality~\cite{du2021compressed}.

\begin{figure}[t]
    \centering
    \includegraphics[width=\columnwidth]{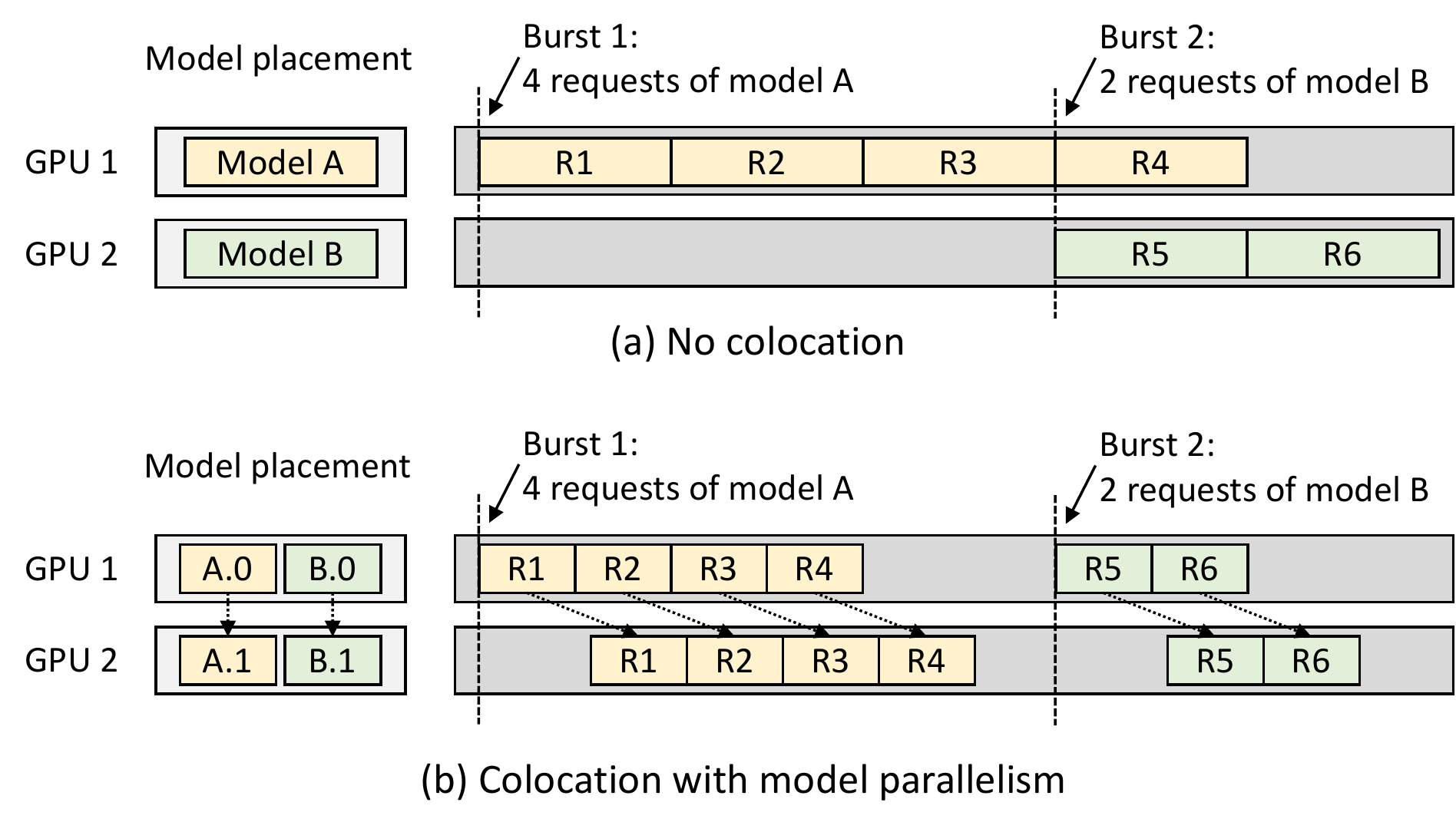}
	\vskip -1em
    \caption{Two placement strategies for serving two models on two GPUs. In each subfigure, the left part shows the model placements and the right part shows the timeline for handling bursty requests. At the time of "Burst 1", 4 requests of model A come at the same time. Colocation with model parallelism can reduce the average completion time of bursty requests.}
    \vskip -1em
    \label{fig:two-model-example}
\end{figure}

Provisioning sufficient resources to serve these models can be arduous as request rates are bursty.
For example, using common workload traces, we observe frequent spikes in demand of up to 50$\times$ the average~\cite{zhang2021faster}.
Meeting the service level objective (SLO) of latency usually means provisioning for these peak loads, which can be very expensive;
additional devices allocated for this purpose would remain underutilized most of the time.
Making matters worse, it is increasingly common to serve multiple models and multiple variations of the same large model
in situations 
like A/B testing or serving fine-tuned models for specific domains (\S\ref{sec:serving-sys-background}).

This paper studies how to efficiently serve multiple large models concurrently.
Specifically, we explore the underappreciated benefits of model parallelism in online model serving, even for smaller models that can fit on a single device.
Model parallelism refers to partitioning and executing a model on distributed devices (\S\ref{subsec:model-parallel-background}). 
The benefits of model parallelism have been well studied~\cite{zheng2022alpa,huang2019gpipe,lepikhin2020gshard,narayanan2021megatron} in the \emph{throughput-oriented} training setting.
However, its effects for model serving under \emph{latency-sensitive} settings remains largely untapped.

We observe that there are fundamental transition points in the model serving design space that challenge prior assumptions about serving, even for models that fit on a single device.
For example, consider the scenario with two models and two GPUs, each of which has sufficient memory to hold one complete model.
As shown in Fig.~\ref{fig:two-model-example}(a), the natural approach assumed by almost all existing serving systems~\cite{nvidiatriton,crankshaw2017clipper,olston2017tensorflow} is to allocate one dedicated GPU for one model.
This approach appears rational because partitioning the model across GPUs would incur communication overheads that would likely increase the prediction latency.
However, we find that inducing additional model parallelism (to the point where per-example execution time actually \emph{increases}) enables a wider range of placement strategies, e.g., model co-location, which can improve the statistical multiplexing of the system under bursty workloads. In Fig.~\ref{fig:two-model-example}(a), assuming the execution time of a model is $y$, the average end-to-end latency of request 1 through 4 is $(1y + 2y + 3y + 4y)/4 =2.5y$. In Fig.~\ref{fig:two-model-example}(b), assuming a 10\% model-parallel overhead, the average latency of request 1 through 4 is reduced to $(1.1y + 1.6y + 2.1y + 2.6y)/4= 1.85y$. 
Co-location with model parallelism can utilize more devices to handle bursty requests and reduces the average completion time, despite its overheads (\S\ref{subsec:illustrative-example}). Even if we batch the requests, the case still holds (\S\ref{subsec:batching-exp}).

Unfortunately, the decision of how to optimally split and place a collection of models is complex.
Although leveraging model parallelism as above has its benefits, it still adds overheads that may negate those benefits for less bursty workloads.
For example, we find that a particularly influential axis on the efficacy of model parallelism is per-GPU memory capacity (\S\ref{subsec:serving-performance-factors}), although other factors (e.g., the arrival pattern, SLO) can also have a significant effect.
Further, besides the inter-op model parallelism presented in \cref{fig:two-model-example}, another kind of model parallelism, intra-op parallelism, presents its own distinct tradeoffs (\S\ref{subsec:model-parallel-overhead}).
Ultimately, different styles of parallelism and their tradeoffs create a complex, multi-dimensional, and multi-objective design space that existing systems largely ignore and/or fail to navigate.
However, not leveraging model parallelism in the serving setting is typically not an option for large models, and not addressing this trade-off space directly results in significant increases in cost and serving latency.

To that end, we present \sys\footnote{https://github.com/alpa-projects/mms}, a system that automatically and efficiently explores the tradeoffs among different parallelization and placement strategies for model serving.
\sys takes a cluster resource specification, a set of models, and a periodic workload profile; it then partitions and places the models and schedules the requests to optimize SLO attainment (i.e., the percentage of requests served within SLO).
To assist the design of \sys, we first introduce a taxonomy and quantify the tradeoffs between different parallelization strategies in model serving (\S\ref{sec:motivation}).
We then present key algorithms to navigate the tradeoff space (\S\ref{sec:methods}).
We design an iterative simulator-guided model placement algorithm to optimize the colocation of models and a group partition algorithm to search for the best way to partition the cluster into disjoint model-parallel groups.
In addition, we extend the existing auto-parallelization algorithms for training to make them more suitable for inference.

We evaluate \sys with production workloads on a 64-GPU cluster (\S\ref{sec:evaluation}).
Evaluation results show that, when optimizing one metric at a time, \sys can choose to increase the request processing rate by 10$\times$, achieve 2.5$\times$ lower latency deadlines,
or tolerate 6$\times$ burstier traffic compared to previous state-of-the-art serving systems.

In summary, we make the following contributions:

\begin{itemize}
\item A detailed analysis of the tradeoff space of different model parallel strategies for efficient model serving.
\item Novel model placement algorithms to incorporate model parallelism in a serving system. 
\item A comprehensive evaluation of \sys with both synthetic and production workloads.
\end{itemize}

%% file: sec.background.tex
\section{Background}
\label{sec:serving-sys-background}

Over the past few years, increasingly capable models have been developed for everything from recommendations to text generation. As a result, serving predictions from these models has become an essential workload in modern cloud systems.
The structure of these workloads often follows a simple request-response paradigm. Developers upload a pre-trained model and its weights ahead of time;
at runtime, clients (either users or other applications) submit requests for that model to a serving system, which will queue the requests, dispatch them to available GPUs/TPUs, and return the results.

The requirements of these model-serving systems can be stringent.
To satisfy user demand, systems often must adhere to aggressive SLO on latency.
At the same time, serving systems that must run continuously need to minimize their operational costs associated with expensive accelerators.
Minimizing serving costs can be challenging because dynamically scaling compute resources would be too slow on the critical path of each prediction request: it can take multiple seconds just to swap a large model into accelerator memory~\cite{rajbhandari2020zero}. 
Furthermore, there is significant and unpredictable burstiness in the arrival process of user requests.
To meet tight SLO, contemporary serving systems are forced to over-provision compute resources, resulting in low cluster utilization~\cite{weng2022mlaas}.

Another pattern that emerges in serving large models is the use of multiple instances of the same or similar model architectures.
This is commonly seen in the practice of pretraining on large unlabeled data and fine-tuning for various downstream tasks~\cite{devlin2018bert}, which can significantly boost accuracy but results in multiple instances of the same model architecture.
For example, Hugging Face serves more than 9,000 versions of fine-tuned BERT~\cite{huggginface}.
They either share a portion of the parameters or do not share any parameters at all for better accuracy.
Prior works have ~\cite{zhou2022pets,shen2019nexus} exploited the property of shared parameters, but we do not consider the shared parameters in this paper because \sys targets general settings and full-weight tuning is still a major use case.

\begin{figure*}
    \centering
     \begin{subfigure}[b]{0.24\textwidth}
     \centering
     \includegraphics[width=\textwidth]{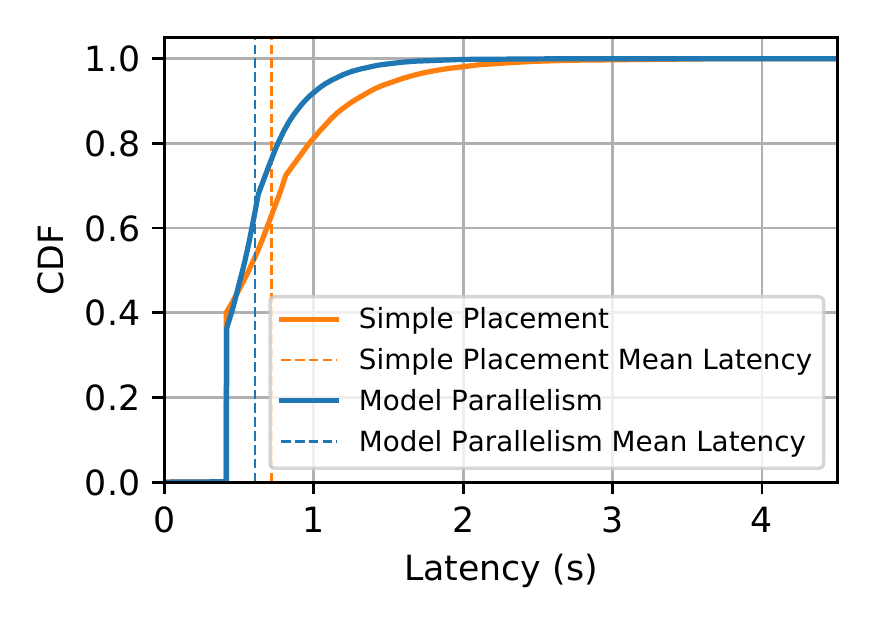}
     \vspace{-6mm} \caption{Poisson arrival.}
     \label{fig:illustrative-example-uniform}
    \end{subfigure}
     \begin{subfigure}[b]{0.24\textwidth}
     \centering
     \includegraphics[width=\textwidth]{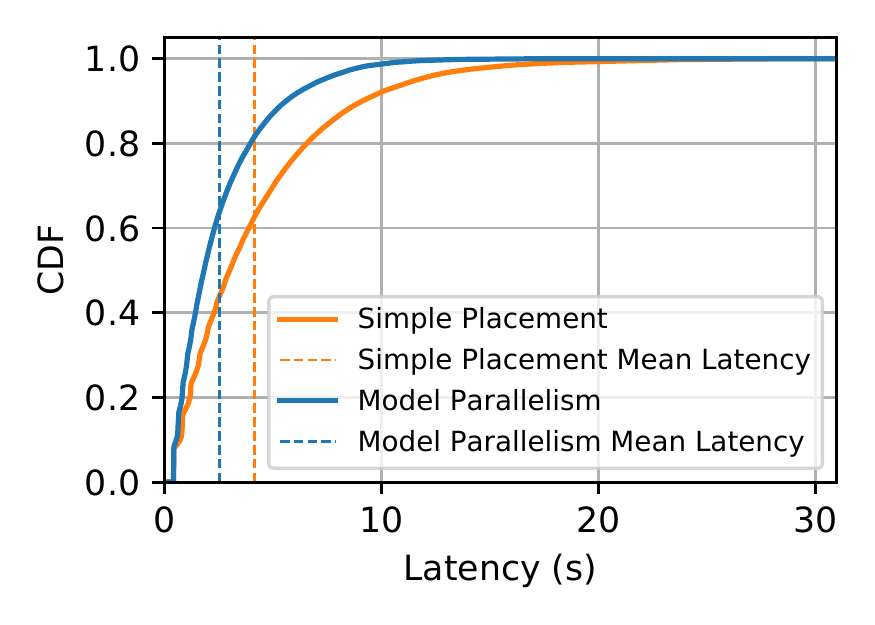}
     \vspace{-6mm}\caption{High CV Gamma arrival.}
     \label{fig:illustrative-example-high-cv}
    \end{subfigure}
     \begin{subfigure}[b]{0.24\textwidth}
     \centering
     \includegraphics[width=\textwidth]{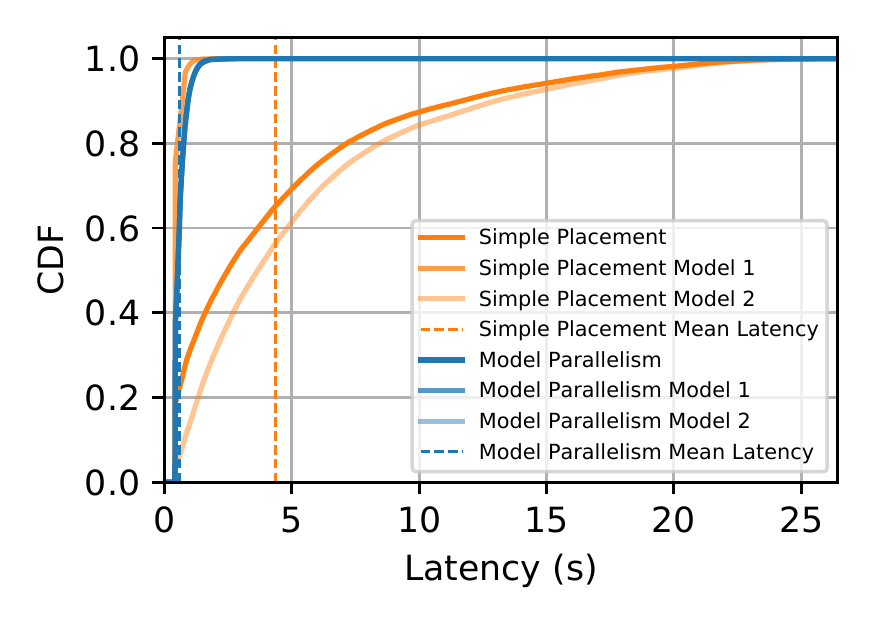}
     \vspace{-6mm}\caption{Different rates.}
     \label{fig:illustrative-example-nonuniform}
    \end{subfigure}
     \begin{subfigure}[b]{0.24\textwidth}
     \centering
     \includegraphics[width=\textwidth]{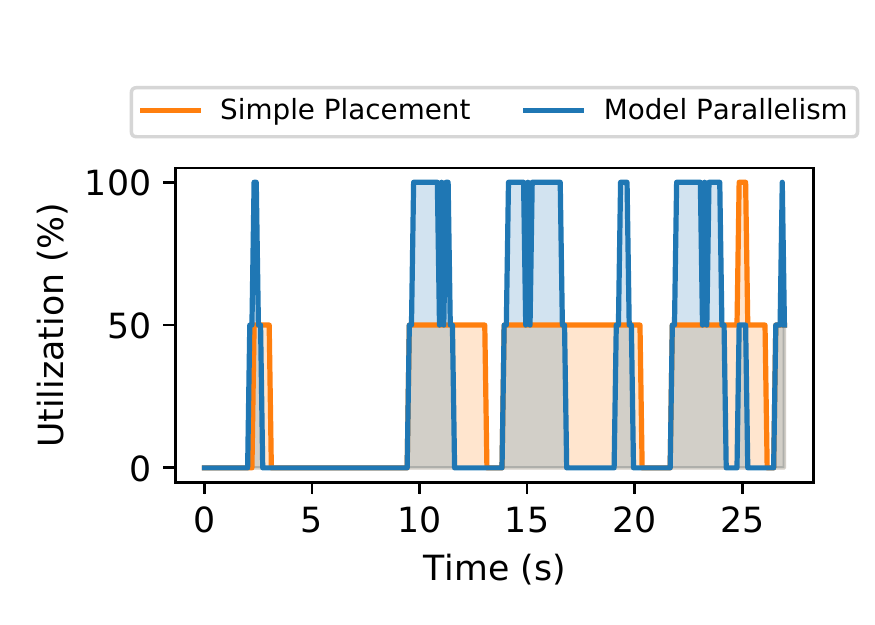}
     \vspace{-6mm}\caption{Cluster utilization.}
    \label{fig:cluster-utilization-illustration}
    \end{subfigure}
    \vspace{-2mm}
    \caption{Latency CDF and cluster utilization in the 2-model example.}\vspace{-2mm}
    \label{fig:illustrative-example}
\end{figure*}

\subsection{Model Parallelism in Model Serving}
\label{subsec:model-parallel-background}

Distributed parallel model execution is necessary when attempting to satisfy the serving performance requirements or support large models that do not fit in the memory of a single device. 

At a high level, distributed execution of deep learning models can be classified into two categories: intra-operator parallelism
and inter-operator parallelism~\cite{zheng2022alpa}.

\heading{Intra-operator parallelism.}
DL models are composed of a series of operators over multidimensional tensors, e.g., matrix multiplication over input and weight tensors.
Intra-operator parallelism is when a single operator is partitioned across multiple devices, with each device executing a portion of the computation in parallel \cite{xu2021gspmd, shazeer2018mesh, shoeybi2019megatronv1}.
Depending on the specific partitioning strategy and its relationship to prior and subsequent operators in the model, partitioning can require communication among participating GPUs to split the input and then merge the output.

The benefit of intra-operator parallelism for single-request execution is twofold.
First, it can expand the total amount of computation available to the target model, reducing its end-to-end latency.
In a similar fashion, it can expand the total memory available to the model for storing its inputs, weights, and intermediate values.
The cost is the aforementioned communication overhead.

\heading{Inter-operator parallelism.}
The other type of parallelism available to DL models is inter-operator parallelism, which assigns different operators of the model's execution graph to execute on distributed devices in a pipeline fashion (a.k.a. pipeline parallelism) \cite{huang2019gpipe, narayanan2019pipedream, li2021terapipe}.
Here, devices communicate only between pipeline stages, typically using point-to-point communication between device pairs.

Unlike intra-operator parallelism, pipeline parallelism does not reduce the execution time of a single request.
In fact, it typically increases the execution time due to modest amounts of communication latency between pipeline stages, although the total amount of transferred data is often lower than it is in intra-operator parallelism.
Instead, the primary use of inter-operator parallelism 
in traditional serving systems 
is to allow the model to exceed the memory limitation of a single GPU.

%% file: sec.motivation.tex
\section{Motivation and Tradeoff Analysis}
\label{sec:motivation}

As mentioned, both types of model parallelism reduce per-device memory usage by partitioning a model on multiple devices.
A key motivation for this work is that we can use this property to fit more models on one device, enabling better statistical multiplexing of the devices when handling bursty requests.
We explore this idea through a series of empirical examinations and theoretical analysis, starting with an illustrative example (\S\ref{subsec:illustrative-example}), followed by an empirical analysis of when model parallelism is beneficial (\S\ref{subsec:serving-performance-factors}), the overhead of model parallelism (\S\ref{subsec:model-parallel-overhead}), and a queueing theory-based analysis (\S\ref{subsec:queueing-theory-analysis}).
All the experiments in this section are performed on an AWS EC2 p3.16xlarge instance with 8 NVIDIA 16GB V100 GPUs.

\subsection{Case Study: A Two-model Example}

\label{subsec:illustrative-example}

We start with an illustrative experiment to show how model parallelism can benefit the serving of multiple models.
We use two GPUs to serve two Transformer models with 6.7 billion parameters each (13.4 GB to store its FP16 weights).
Because each GPU has 16\,GB of memory, it can fit one and only one model. A single request takes around 0.4\,s to process on one GPU.

We compare the following model placements, corresponding to the strategies in \cref{fig:two-model-example}.
The first is \emph{simple placement}, where we place one model on each GPU due to the memory constraint. 
The second is \emph{model-parallel placement}, where we use inter-op parallelism to partition each model to a 2-stage pipeline and let each GPU execute half of each model.

We evaluate the two placements when the requests to each model follow an independent Poisson process with an arrival rate of 1.5 request/s.
\cref{fig:illustrative-example-uniform} shows the cumulative distribution function (CDF) and average of request latency (which includes the GPU execution time and queuing delay).
Model-parallel placement reduces the average latency of the simple placement from 0.70s to 0.55s, a 1.3$\times$ speedup.
The speedup comes from the better burst tolerance: when a burst arrives that exceeds the capability of a single GPU, simple placement must begin queuing requests.
However, as long as the other model does not receive many requests, the model parallel placement can use both GPUs to serve the requests for the popular model via statistical multiplexing of the GPUs. 

This effect becomes more pronounced with higher burstiness, which we can demonstrate using a Gamma request arrival process with the same average request rate as above but a higher coefficient of variance (CV) of 3.
As shown in \cref{fig:illustrative-example-high-cv}, the speedup on mean latency is now increased to 1.9$\times$.
\cref{fig:cluster-utilization-illustration} shows a representative trace of the corresponding total cluster utilization over time.
Note that for each request burst, model-parallel placement can use the whole cluster and only take half of the time to process, while simple placement can only use half of the cluster.

In addition, we also evaluate the case where one model receives more requests than another. In \cref{fig:illustrative-example-nonuniform}, we use Poisson arrival but let 20\% of the requests ask for model 1 and 80\% ask for model 2. Although replication performs slightly better for model 1 requests, it is drastically worse on model 2 requests compared to the model-parallel placement. For model-parallel placement, because both GPUs are shared across two models, the requests to both models follow the same latency distribution.
Overall, model-parallel placement reduces the mean latency by 6.6$\times$.

\subsection{When is Model Parallelism Beneficial}
\label{subsec:serving-performance-factors}

To further explore the nuances of model parallelism in serving, we increase the size of the deployment to 8 GPUs and 8 Transformer models with 2.6B parameters each.
As a base setting, we set the requests to each model as a Gamma process with an average rate of 20 request/s and CV of 3;
we then vary a range of factors to see their effects.
Note that some of the settings we evaluate are impossible on real hardware (e.g., exceeding the memory capacity of a single device) so we leverage the simulator introduced in \S\ref{sec:implementation}. The fidelity of the simulator is very high as verified in \S\ref{subsec:exp-setup}.

\begin{figure}[t]
    \centering
    \vspace{-1mm}
    \includegraphics[width=\columnwidth]{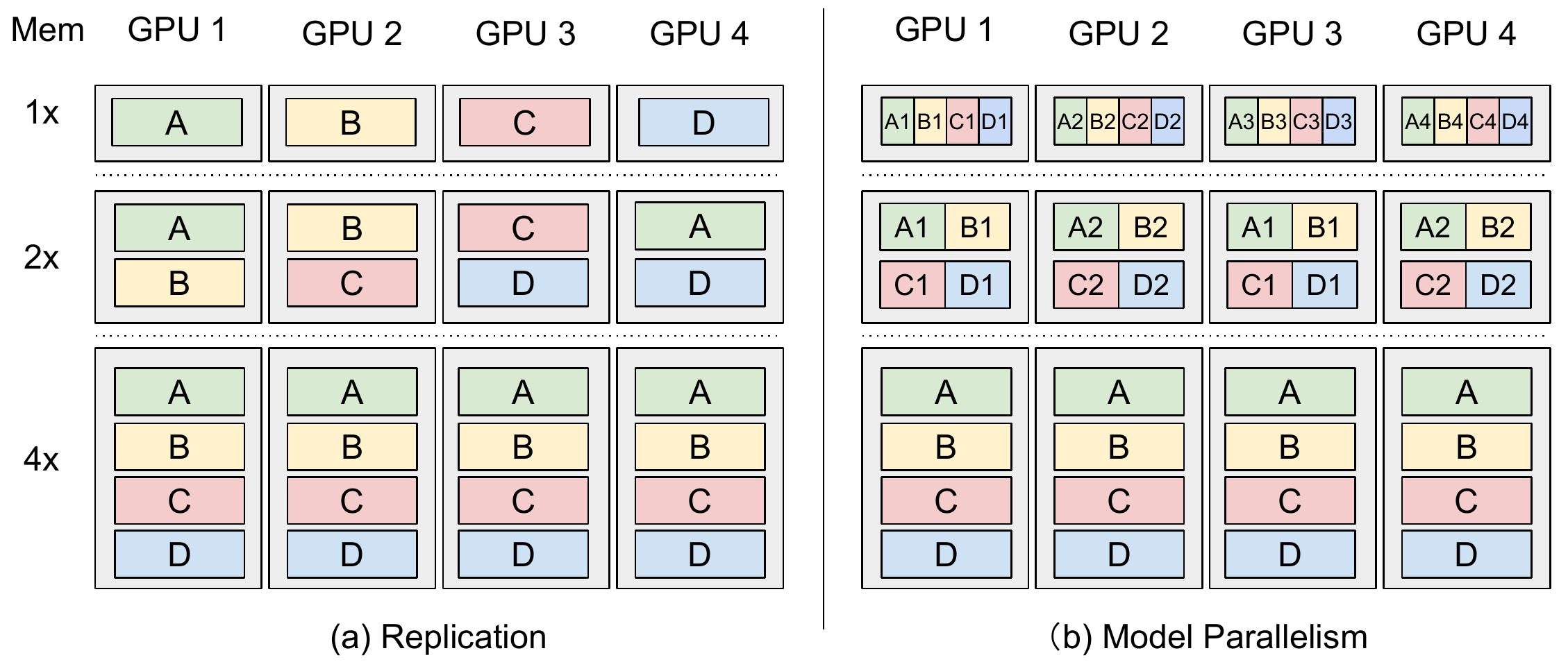}
    \vspace{-5mm}
    \caption{Replication and model parallel placement illustration with different memory budgets, where the memory budgets are set to be multiples of a single model's size.}
    \vspace{-2mm}
    \label{fig:selective-replication-mem-illustration}
\end{figure}

\begin{figure}
    \centering
     \includegraphics[width=.23\textwidth]{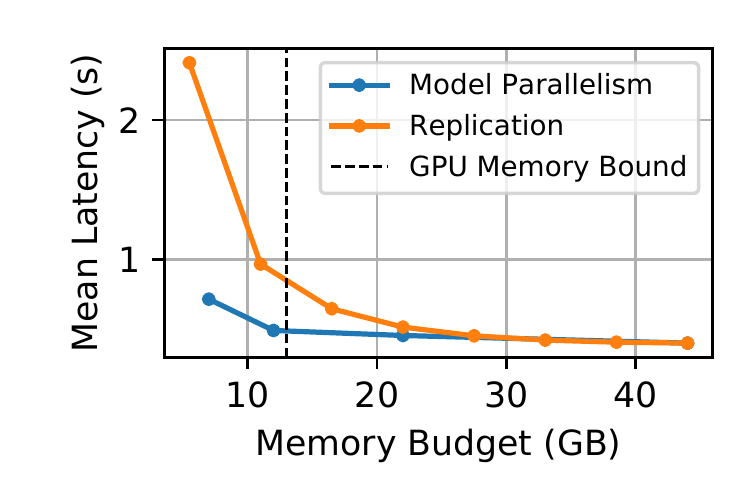}
     \includegraphics[width=.23\textwidth]{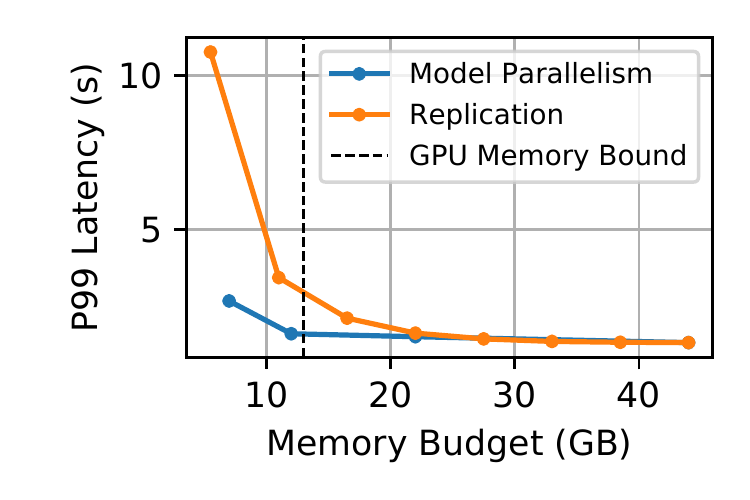}
    \vspace{-2mm}
    \caption{Serving performance with changing per-GPU memory budgets. Model parallelism is beneficial for limited memory budget. The dashed vertical line is the real per-GPU memory bound of a 16GB V100. The value is around 13GB due to the need to store activations and other runtime context. }
    \label{fig:mem-latency}
    \vspace{-4mm}
\end{figure}

The model in this case is smaller (5.2GB), so one GPU can also store multiple models without model parallelism.
We compare two placement methods: (1) \emph{Replication.} In this setting, we replicate the models to different devices until each device cannot hold any extra models. Because all the models receive equal amounts of loads on average, we replicate each model the same number of times (\cref{fig:selective-replication-mem-illustration}a). (2) \emph{Model Parallelism.} Here we use inter-operator parallelism and uniformly assign the Transformer layers to different GPUs.

\begin{figure}
    \centering
     \includegraphics[width=.23\textwidth]{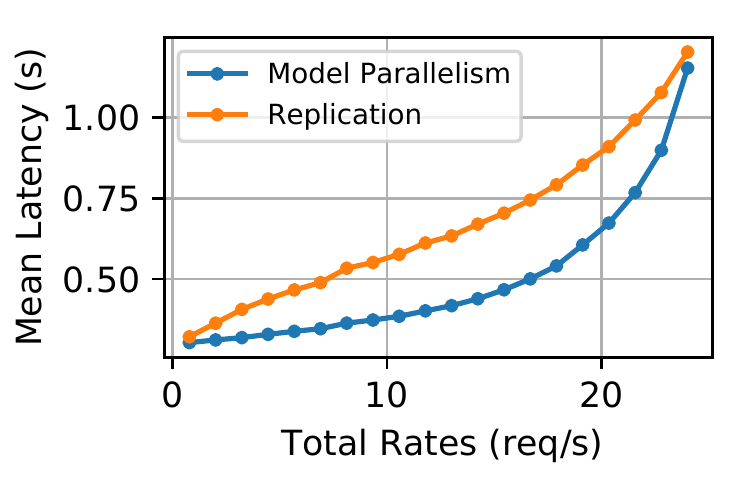}
     \includegraphics[width=.23\textwidth]{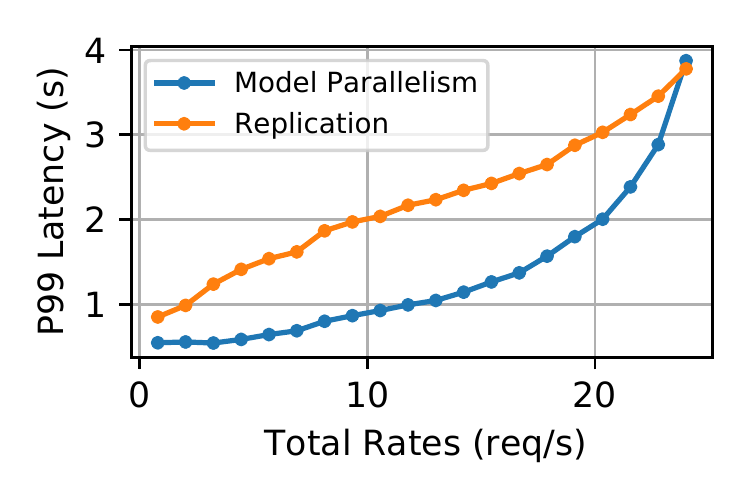}
    \vspace{-2mm}
    \caption{Serving performance with changing arrival rates. Model parallelism is beneficial for smaller rates.}
    \label{fig:changing-rate}
    \vspace{-2mm}
\end{figure}

\heading{Device memory.} We evaluate the mean and the tail latency of the two placement methods under different device memory capacities. For replication, more GPU memory can fit more models onto a single GPU. For model parallelism, more GPU memory can also reduce the number of pipeline stages and reduce the overhead as in \cref{fig:selective-replication-mem-illustration}b. The resulting mean and P99 latency is shown in \cref{fig:mem-latency}. With more memory, more models can fit into a single GPU, so the benefit of statistical multiplexing diminishes because replication can also effectively use multiple devices to serve the bursty requests to a single model. When the GPU memory capacity is large enough to hold all models, there is no gain from model parallelism.

\begin{figure}[t]
    \centering
     \includegraphics[width=.23\textwidth]{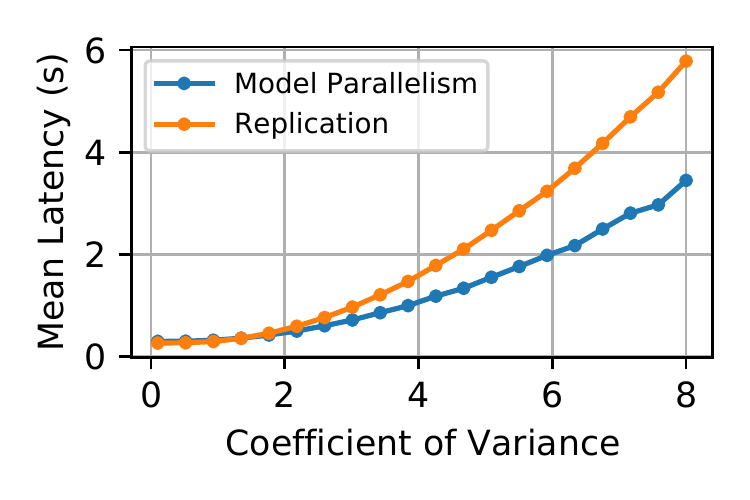}
     \includegraphics[width=.23\textwidth]{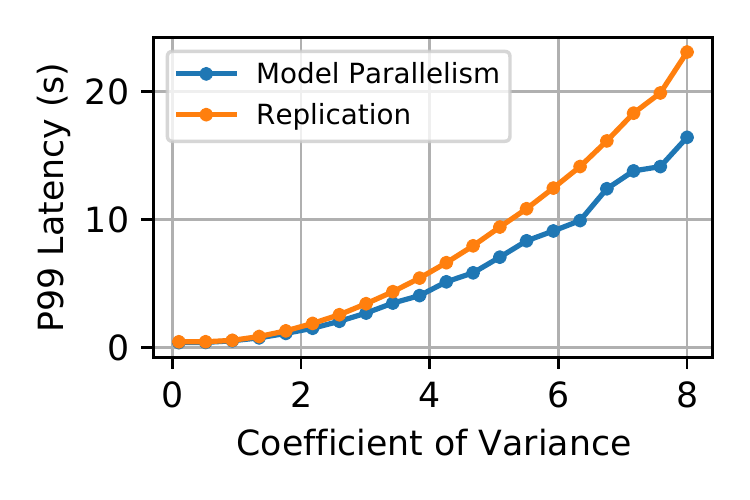}
    \vspace{-2mm}
    \caption{Serving performance with changing CVs. Model parallelism is beneficial for larger CVs.}
    \label{fig:changing-cv}
    \vspace{-4mm}
\end{figure}

\heading{Request arrival.} We vary the parameters of the arrival process
and compare the replication placement with the model-parallel placement with 8-stage pipeline parallelism.
The mean and P99 latency results of changing arrival rate are shown in \cref{fig:changing-rate}. When the arrival rate is low, model parallelism can greatly reduce the serving latency. However, when the arrival rate approaches the peak serving rate of the cluster, the benefit of model-parallel placement starts to diminish. Eventually, it starts to perform worse than replication.
This is because when all models are equally saturated, the replication placement is able to achieve efficient cluster utilization and there is no benefit to the statistical multiplexing afforded by model parallelism.
Instead, the overhead of model parallelism (\S\ref{subsec:model-parallel-overhead}) starts to become a significant factor.

The mean and P99 latency results of changing CV are in \cref{fig:changing-cv}. With a higher CV, the requests become more bursty, and the benefit of model parallelism becomes more significant. As shown in the results, with a higher CV, model parallelism can greatly outperform the performance of replication.

\begin{figure}[t]
    \vspace{-4mm}
    \centering
     \begin{subfigure}[t]{0.23\textwidth}
     \centering
     \includegraphics[width=\textwidth]{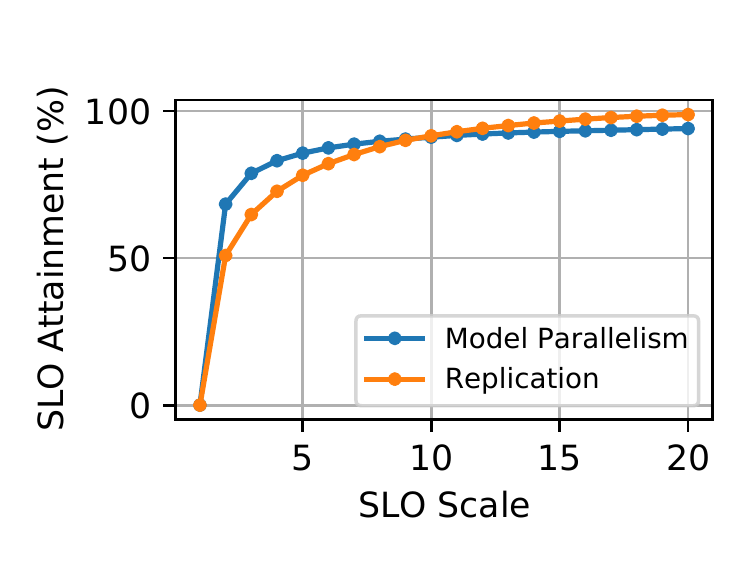}
     \vspace{-7mm}
     \caption{Real model latency.}
     \label{fig:changing-slo-goodput}
    \end{subfigure}
     \begin{subfigure}[t]{0.23\textwidth}
     \centering
     \includegraphics[width=\textwidth]{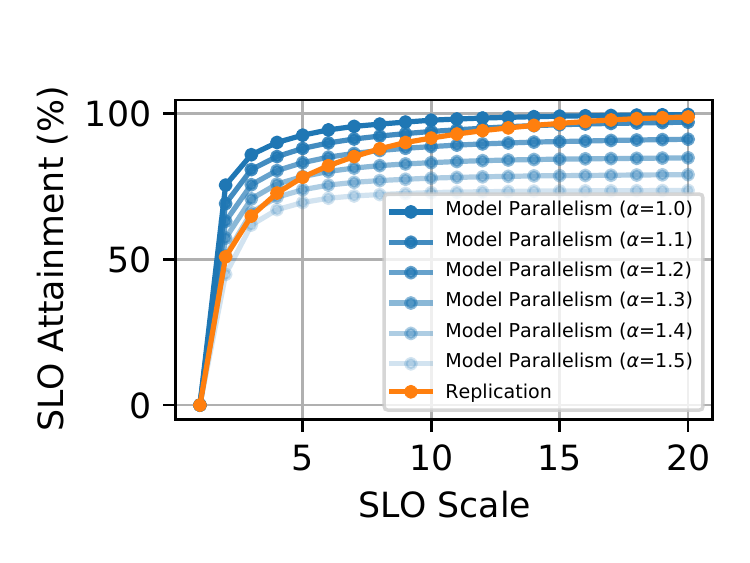}
     \vspace{-7mm}
     \caption{Changing overhead.}
     \label{fig:changing-slo-goodput-custom}
    \end{subfigure}
    \vspace{-2mm}
    \caption{SLO attainment with changing SLOs. Model parallelism is beneficial for smaller SLOs.}
    \vspace{-5mm}

\end{figure}

\paragraph{Service level objectives.} In prediction serving settings, it is common to have tight latency SLO and predictions made after these deadlines are often discarded \cite{gujarati2020serving}. 
For example, advertising systems may choose not to show an ad rather than delay rendering user content. 
In this case, the goal of the serving system is to optimize the percentage of requests that can be finished within the deadline, i.e., \emph{SLO attainment}.

In this experiment, we measure how SLOs affect the performance of the placement methods. We compare the replication and the model-parallel placement with 8-stage pipeline parallelism. During execution, we drop the requests that will exceed the deadline even if we schedule it immediately. We scale the SLO to different multiplies of the single device execution latency (\emph{SLO Scale} in \cref{fig:changing-slo-goodput}) and compare the SLO attainment of the two methods.

As in \cref{fig:changing-slo-goodput}, when SLO is tight ($<10\times$ model latency), model parallelism can greatly improve SLO attainment. However, when the SLO becomes looser, its SLO attainment plateaus but that of the replication placement keeps growing. This result shares the same core logic as previous experiments: When SLO becomes looser, more requests can stay in the waiting queue, and thus the effective burstiness of the requests decreases. When many requests are queued, the system is bounded by its total processing capability, which might be affected by the model parallelism overhead. In the real world, the SLO requirement is often less than $5\times$ of the model execution latency \cite{gujarati2020serving}, where model parallelism can improve SLO attainment.

\vspace{1mm}
\noindent\hspace{0.02\linewidth}\fbox{\begin{minipage}{0.94\linewidth}
\textbf{Summary}: Model parallelism benefits model serving through statistical multiplexing when the device memory is limited, the request rate is low, the request CV is high, or the SLO is tight.
\end{minipage}}

\subsection{Overhead of Model Parallelism}
\label{subsec:model-parallel-overhead}

\begin{figure}
    \centering
     \begin{subfigure}[b]{0.22\textwidth}
     \centering
     \includegraphics[width=\textwidth]{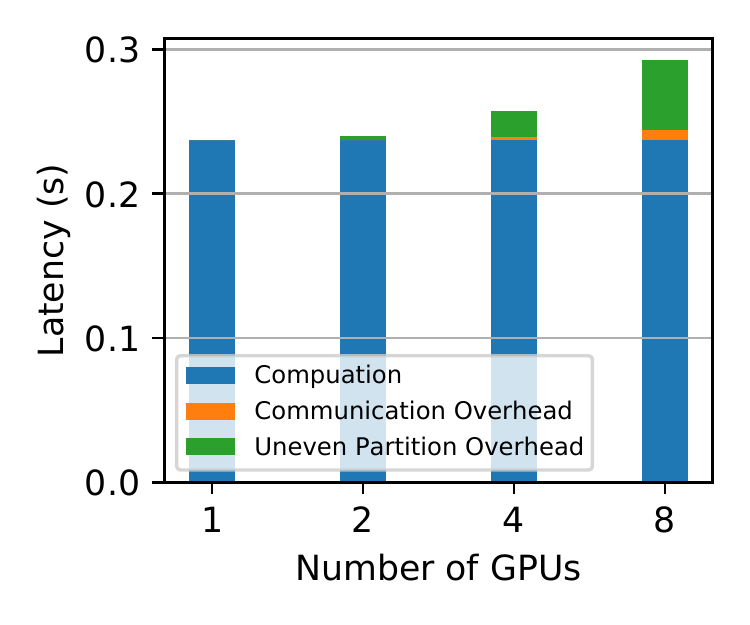}
    \vspace{-6mm}\caption{Inter-op parallelism.}
     \label{fig:overhead-decomposition-pp}
    \end{subfigure}
    \quad
     \begin{subfigure}[b]{0.22\textwidth}
     \centering
     \includegraphics[width=\textwidth]{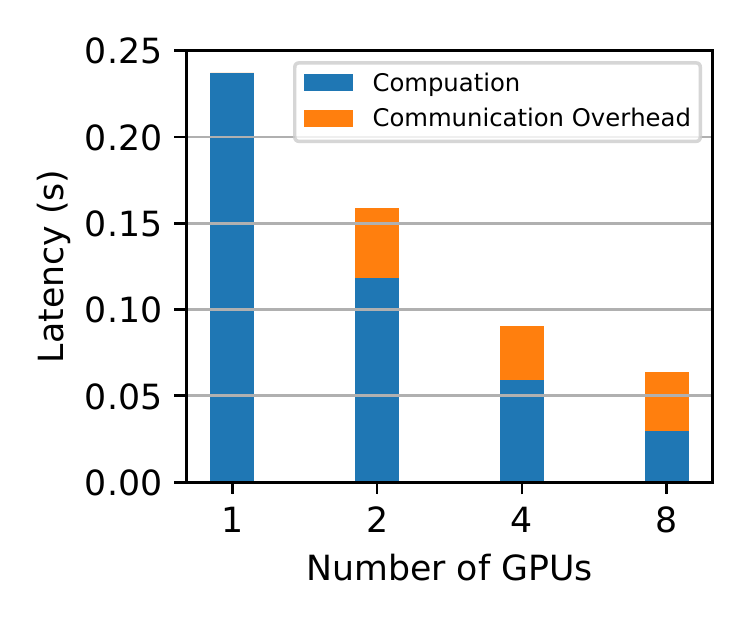}
     \vspace{-6mm}\caption{Intra-op parallelism.}
     \label{fig:overhead-decomposition-op}
    \end{subfigure}
    \vspace{-2mm}
    \caption{The overhead decomposition. The overhead of inter-op parallelism mainly comes from uneven partition while the overhead of intra-op parallelism comes from communication.}
    \vspace{-6mm}
    \label{fig:overhead-decomposition}
\end{figure}

\begin{figure*}
    \centering
     \begin{subfigure}[b]{0.28\textwidth}
     \centering
     \includegraphics[width=\textwidth]{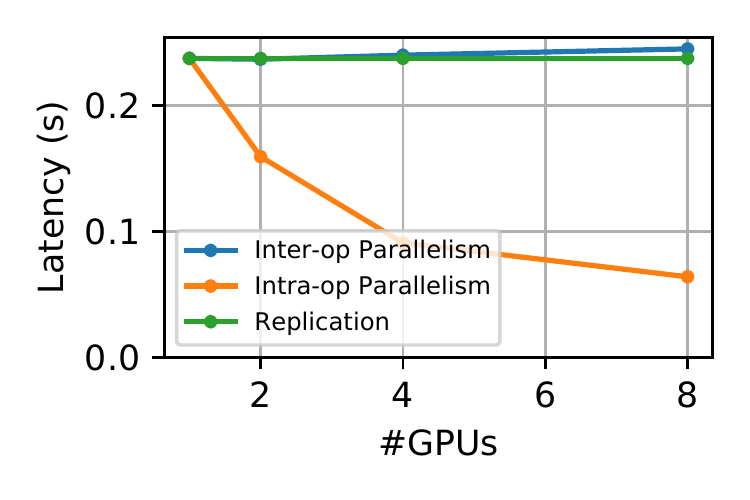}
     \vspace{-6mm}\caption{Single input latency.}
     \label{fig:model-parallel-latency}
    \end{subfigure}
    \quad
     \begin{subfigure}[b]{0.28\textwidth}
     \centering
     \includegraphics[width=\textwidth]{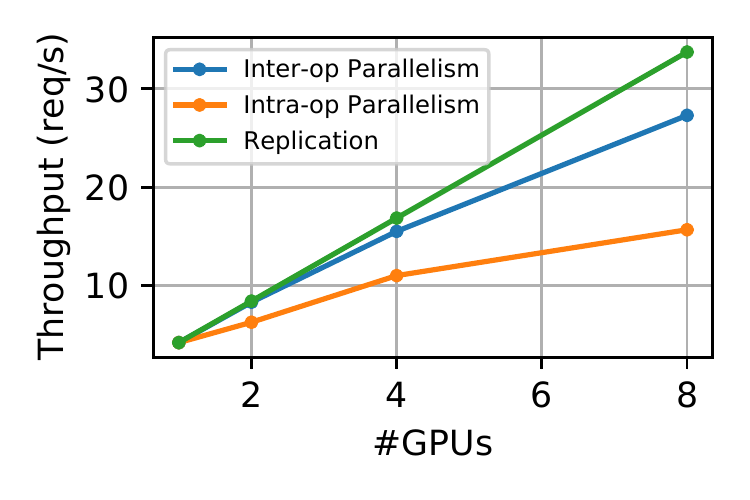}
     \vspace{-6mm}\caption{Throughput.}
     \label{fig:model-parallel-throughput}
    \end{subfigure}
    \quad
     \begin{subfigure}[b]{0.28\textwidth}
     \centering
     \includegraphics[width=\textwidth]{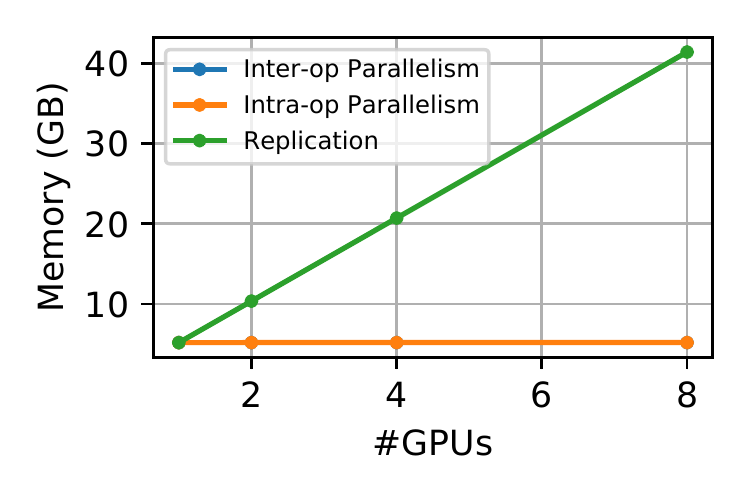}
     \vspace{-6mm}\caption{Total memory usage.}
     \label{fig:model-parallel-memory}
    \end{subfigure}

    \caption{The latency, throughput and memory usage vs. \#GPUs for inter-op parallelism, intra-op parallelism, and replication. In subfigure (c), the lines for inter-op and intra-op parallelism overlap.}
    \label{fig:model-parallel-throughput-latency-memory}
\end{figure*}

In this section, we further investigate the overheads of different model parallel strategies and how they affect serving performance.
Similar to the setup in \cref{fig:changing-slo-goodput}, we manually modify the overhead of model parallelism. 
Specifically, let the latency of a single model executing on the GPU be $L$ and the number of pipeline stages be $n$. We set the total latency of pipeline execution to be $\alpha L$ and the latency of each pipeline stage to be $\alpha L/n,$  where $\alpha$ is a parameter that controls the overhead. When $\alpha = 1,$ model parallelism does not have any overhead and larger $\alpha$ means higher overhead.

We show the results in \cref{fig:changing-slo-goodput-custom}. If model parallelism does not have any overhead ($\alpha = 1$), it can always outperform replication due to its ability to multiplex the devices. 
When the overhead becomes larger and the SLO is low, model parallelism still outperforms replication. 
However, with a larger SLO, the effective burstiness is reduced and the performance is dominated by the overhead.

Given that the overhead can greatly affect serving performance, we perform a detailed study of the multiple sources of
model-parallel overhead in \cref{fig:overhead-decomposition}. For inter-op parallelism, when partitioning a single model into multiple stages, different stages need to communicate the intermediate tensors, and we denote this overhead as the \emph{communication overhead}. In addition, the pipeline execution will be bottlenecked by the execution time of the slowest stage, making the effective latency to be the number of pipeline stages times the latency of the slowest stage \cite{huang2019gpipe}. We denote this as the \emph{uneven partition overhead}. As in \cref{fig:overhead-decomposition-pp}, for inter-op parallelism, most overhead comes from the latency imbalance among different pipeline stages, instead of the communication between stages. While our previous discussion mainly focuses on inter-op parallelism, the other type of model parallelism, intra-op parallelism, has very different performance characteristics. Its overhead is merely brought by the collective communication across multiple devices \cite{narayanan2021megatron}, which cannot be overlapped with the neural network computation due to data dependency. From \cref{fig:overhead-decomposition-op}, we can see that the communication overhead of intra-op parallelism is much higher than inter-op parallelism.

Finally, we compare the latency, throughput, and memory consumption of different model-parallel placements and the replication placement in \cref{fig:model-parallel-throughput-latency-memory}. Because of the sequential dependence between the different stages, inter-op parallelism cannot reduce the execution latency of a single input data. Instead, the latency is slightly higher due to the communication between the stages. On the other hand, intra-op parallelism can largely reduce the latency via the parallel execution of different GPUs (\cref{fig:model-parallel-latency}). However, because inter-op parallelism can pipeline the execution of different stages and only communicate a relatively small amount of data, it attains higher throughput compared to intra-op parallelism (\cref{fig:model-parallel-throughput}). Because both parallel methods split the model weight tensors across different GPUs, the total memory usage stays constant with increasing numbers of GPUs (\cref{fig:model-parallel-memory}). This makes the statistical multiplexing of different GPUs across multiple models possible.

In the end, the tradeoff between parallelization strategies and their interplay with cluster resources, arrival patterns, and serving objectives forms an intricate design space.

\subsection{Queueing Theory Analysis}
\label{subsec:queueing-theory-analysis}

In this section, we use queuing theory to mathematically verify the conclusions in \S\ref{subsec:serving-performance-factors} and \S\ref{subsec:model-parallel-overhead}. Specifically, we analyze the case where the inputs follow the Poisson arrival process. Since the execution time of a deep learning inference task is highly predictable \cite{gujarati2020serving}, we assume the request serving time is deterministic. For the single device case, suppose the request rate to a model is $\lambda_0$ and the single device latency is $D$ with the utilization $\lambda_0 D < 1$, then the average number of requests $L_Q$ and the average latency $W$ in this M/D/1 queue \cite{shortle2018fundamentals} are:
\[
L_Q = \frac{\lambda_0 D}{2 (1 - \lambda_0 D)},\quad W = D + L_Q D = D + \frac{\lambda_0 D^2}{2 (1 - \lambda_0 D)}.
\]

Now consider the example in \S\ref{subsec:illustrative-example}. Let $p\lambda, (1 - p)\lambda$ be the average request rates for the two models respectively, where $p \in [0, 1]$ controls the percentage of requests for both models. Then for the simple placement, the average latency can be derived as the average latency of two independent queues:
\[
W_\mathit{simple} = D + \frac{p^2\lambda D^2}{2 (1 - p\lambda D)} + \frac{(1 - p)^2\lambda D^2}{2 (1 - (1 - p)\lambda D)}.
\]
Note that $W_\mathit{simple}$ reaches minimum when $p = 1/2.$ Intuitively, when $p$ is not exactly half, one model receives more requests than the other. This larger portion of requests have a longer queueing delay, which leads to the higher overall mean latency.

For the model-parallel case, the requests to both models merged to a single Poisson Process with rate $\lambda.$ For pipeline parallelism, suppose the latency for a single input to be $D_s$ and the maximum stage latency to be $D_m$, then the average latency would be
\[
W_\mathit{pipeline} = D_s + \frac{\lambda D_m^2}{2 (1 - \lambda D_m)}.
\]

Suppose there is no model-parallel overhead, then $D_s = 2 D_m = D.$ Let's first consider the case where $p = 1/2$ (\cref{fig:illustrative-example-uniform}). We have
\begin{align*}
    W_\mathit{simple} = D + \frac{\lambda D^2}{4 - 2\lambda D}, \quad W_\mathit{pipeline} = D + \frac{\lambda D^2}{8 - 4\lambda D}.
\end{align*}
In this case, the waiting time for model-parallel execution is half of the simple placement waiting time, as shown in the vertical lines in \cref{fig:illustrative-example-uniform}. When the $p$ is not $1/2,$ $W_\mathit{simple}$ will increase while $W_\mathit{pipeline}$ will stay the same, so the gap between $W_\mathit{simple}$ and $W_\mathit{pipeline}$ will be even larger, as in \cref{fig:illustrative-example-nonuniform}.

\begin{figure}
    \centering    
    \vspace{-2mm}
    \includegraphics[width=.25\textwidth]{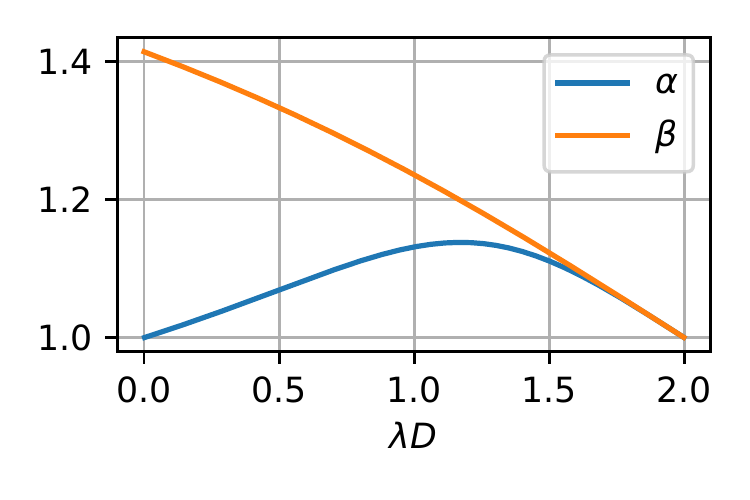}
    \vspace{-5mm}
    \caption{Maximal communication overhead $\alpha$ and uneven partition overhead $\beta$ satisfy $W_\mathit{pipeline} \le W_\mathit{simple}$ as a function of total utilization $\lambda D$.}
    \vspace{-5mm}
    \label{fig:queueing_theory}
\end{figure}

Next, we consider the case where model parallelism incurs overhead. We measure the two types of overheads in \S\ref{subsec:model-parallel-overhead} separately: With the overhead from communication, $D_s = 2D_m = \alpha D$, where $\alpha \ge 1$ is the overhead factor. With the overhead from uneven stages, we suppose $D_s = D$ still holds, but $D_m = \beta D /2$ where $\beta \ge 1$ is the overhead factor. To keep $W_\mathit{pipeline} \le W_\mathit{simple},$ we can get the maximal $\alpha$ and $\beta$ as a function of the total utilization $\lambda D$ separately and we visualize the function in \cref{fig:queueing_theory}. When the utilization is high, the benefit of statistical multiplexing diminishes, and thus the overhead needs to be low, as in \S\ref{subsec:serving-performance-factors}. On the other hand, when the utilization is very low, most requests will not be queued, and thus the communication overhead $\alpha$ needs to be low to keep the processing latency to be low. Note that the maximal overhead here is based on a uniform Poisson arrival distribution. A more bursty or more non-uniform arrival distribution will make the simple placement performs worse and make the model-parallelism placement outperforms the simple replication placement with even higher overhead.

%% file: sec.method.tex
\section{Method}
\label{sec:methods}

From \S\ref{sec:motivation}, we can see that there are several key challenges to effectively utilize model parallelism for deep learning serving:

\begin{itemize}
\item Derive efficient model parallel strategies for inference to reduce the overhead of model parallelism. Specifically, find a partitioning strategy that minimizes the stage imbalance for inter-operator parallelism.
\item Determine model-parallel placements according to the arrival pattern to maximize SLO attainment.
\end{itemize}

\begin{figure}
    \centering
    \includegraphics[width=0.98\columnwidth]{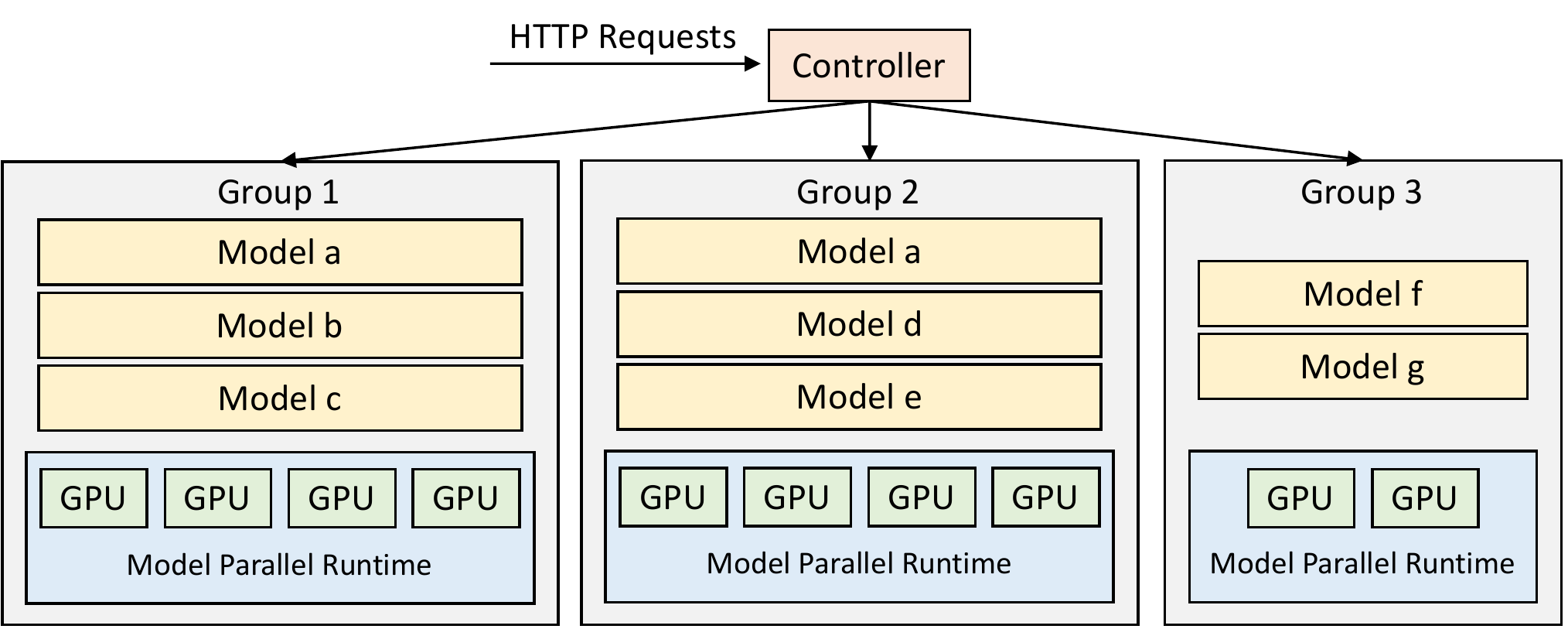}
    \caption{\sys Runtime System Architecture}
    \vspace{-2mm}
    \label{fig:architecture}
\end{figure}

We built \emph{\sys} to specifically tackle these challenges. The runtime architecture of \sys is shown in Fig.~\ref{fig:architecture}. \sys utilizes a centralized controller to dispatch requests to different groups.\footnote{For a larger service, \sys can be extended as a hierarchical deployment with each controller only managing a subset of devices as in \cite{zhang2023shepherd}.} Each group hosts several model replicas on a shared model-parallel runtime. This section describes the architecture of \sys and the key algorithms for efficiently leveraging model parallelism in a model serving system. 

\subsection{Automatic Parallelization for Inference}
\label{subsec:auto-parallel}

Since different parallelization configurations have different latency and throughput trade-offs, we need to enumerate multiple possible configurations for every single model and let the placement algorithm choose the best combination for all models in the whole cluster. Therefore, given a model, \sys first runs an auto-parallelization compiler with various constraints to generate a list of possible configurations. We build several extensions on top of an existing auto parallelization training system, Alpa~\cite{zheng2022alpa}, to make it suitable for generating serving parallelization strategies. Alpa includes two passes for generating efficient model parallel partitions: inter-op pass and intra-op pass. The inter-op pass uses a dynamic programming (DP) algorithm to figure out the optimal inter-op parallel plan, and it calls the intra-op pass for each potential pipeline stage, which is formulated as an integer linear programming (ILP) problem, to profile its latency with the optimal intra-op parallel plan. In \sys, we keep the two compilation passes, but extends both passes for serving.

The inter-op pass in Alpa optimizes the overall pipeline execution latency, which includes the time of forward and backward propagation and weight synchronization. However, in serving workloads, only forward propagation is being executed and there is no need for weight synchronization. Therefore, we reformulate the dynamic programming in \sys to merely focus on minimizing the maximal stage latency. Specifically, denote $F(s, k)$ to be the maximum latency when slicing layers 1 to $k$ into $s$ stages.
We can derive $F$ as
\begin{align*}
    F(s, k)  
    = \min_{1 \le i \le k} \left\{\max\{F(s-1, i - 1), \mathit{latency}(i, k)\}\right\},
\label{eq:dp}
\end{align*}
where $\mathit{latency}(i, k)$ denotes the latency of a stage composes of layer $i$ to $k$. In Alpa, the $\mathit{latency}$ function of all possible $O(K^2)$ combinations is being profiled by the intra-op pass because of the complicated dependency between forward and backward passes. In \sys, because the pipeline stages only perform forward propagation and only communicate intermediate results once between layer boundaries, we can accelerate the profiling by only profiling $K$ layers and letting $\mathit{latency}(i, k)$ to be the sum of the latencies for layer $i$ to $k$. This acceleration enables us to efficiently enumerate different inter- and intra-op device partition setups and generate a list of parallel strategies for the placement algorithm in \S\ref{subsec:placement-alg}.

For the intra-op pass, we extend the ILP in Alpa to drop all configurations that use data parallelism. For serving workloads, because there is no need for weight synchronization, data parallelism can be achieved by the replication placement. We leave the decision of whether to replicate a model to the placement algorithm in \S\ref{subsec:placement-alg}.

\subsection{Placement Algorithm}
\label{subsec:placement-alg}

Given a set of models and a fixed cluster, \sys partitions the cluster into several groups of devices. Each group of devices selects a subset of models to serve using a shared model-parallel configuration. Different groups can hold the same model as replicas. The requests for a model are dispatched to the groups with the requested model replica. We call a specific cluster group partition, model selection, and parallel configuration as a \emph{placement}. Our goal is to find a placement that maximizes the SLO attainment.

However, finding the optimal placement is a difficult combinatorial optimization problem. The overall placement configuration space grows exponentially with the number of devices and the number of models. 
To make things worse, the objective ``SLO attainment'' has no simple analytical formula for an arbitrary arrival distribution. Existing tools and approximations from queueing theory can only analyze simple cases in \S\ref{subsec:queueing-theory-analysis} and cannot model more complex situations \cite{shortle2018fundamentals}. Therefore, we resort to a simulator-guided greedy algorithm that calls a simulator to compute SLO attainment.

To compute the SLO attainment with a given set of requests and placement, in \sys, we assume we know the arrival process in advance. Although short-term burstiness is impossible to predict, the arrival pattern over longer timescales (e.g., hours or days) is often predictable~\cite{weng2022mlaas}. Given this predictability, \sys either directly uses the history request traces or fits a distribution from the trace and resamples new traces from the distribution as the input workload to the simulator to compute the SLO attainment.

\begin{algorithm}[t!]
\caption{Simulator-Guided Greedy Model Selection.}\label{alg:greedy-placement}
\begin{algorithmic}
\Require Model list $M$, device group list $G$, group parallel configurations $P$, workload $W$, beam size $k$ (default = 1).
\Ensure The model selection $\mathit{best\_sel}.$ 
\State $\mathit{best\_sel} \leftarrow \emptyset$ 
\State $\mathit{beam\_sels} \leftarrow \{\emptyset\}$
\While{\textbf{true}}
    \State $\mathit{new\_sels} \leftarrow \emptyset$
    \For{$\mathit{sel} \in \mathit{beam\_sels}$}
        \For{$(m, (g, p)) \in M \times (G, P)$}
            \State \textit{// Parallelize the model as in \S\ref{subsec:auto-parallel}.}
            \State $\mathit{m_{parallelized}} \leftarrow \text{parallelize}(m, g, p)$
            \State $\mathit{sel}' \leftarrow
            \mathit{sel}.\text{add\_model\_to\_group}(\mathit{m_{parallelized}}, g)$
            \If{$\mathit{sel}'$ is in memory constraint}
                \State $\mathit{sel}'.\mathit{slo\_attainment} \leftarrow \text{simulate}(\mathit{sel}', W)$
                \State $\mathit{new\_sels}.\text{append}(\mathit{sel}')$
            \EndIf
        \EndFor
    \EndFor
    \If{$\mathit{new\_sels} = \emptyset$}
        \State \textbf{break}
    \EndIf
    \State $\mathit{beam\_sels} \leftarrow \text{top-}k\text{\_SLO\_attainment}(new\_sels)$
    \State $\mathit{sel}^* \leftarrow \text{pick\_highest\_SLO\_attainment}(\mathit{beam\_sels})$
    \If{$\mathit{sel}^*.\mathit{slo\_att}$ > $\mathit{best\_sel}.\mathit{slo\_att}$}
        \State $\mathit{best\_sel} \leftarrow \mathit{sel}^*$
    \EndIf
\EndWhile
\State \textbf{return} $\mathit{best\_sel}$
\end{algorithmic}
\end{algorithm}

We design a two-level placement algorithm: Given a cluster group partition and a shared model-parallel configuration for each group, \cref{alg:greedy-placement} uses a simulator-guided greedy algorithm to decide which models to select for each group. Then, \cref{alg:group-partition} enumerates various potential cluster partitions and parallel configurations and compares the SLO attainment from \cref{alg:greedy-placement} to determine the optimal placement.

Given a cluster group partition with a fixed model-parallel configuration for each group, \cref{alg:greedy-placement} selects model replicas iteratively as a beam search algorithm: At each iteration, it enumerates all possible (model, group) pairs, parallelizes the model on the device group with the algorithms in \S\ref{subsec:auto-parallel}, and checks whether the model can be put on the group under the memory constraint.
For all valid selections, it runs the simulator and computes SLO attainment. It then picks the top-$k$ solutions and enters the next iteration. The algorithm terminates when no more replicas can be put into any groups.

The complexity of \cref{alg:greedy-placement} is $O(MGRSB)$, where $M$ is the number of models, $G$ is the number of groups, $R$ is the number of replicas we can put according to the memory constraint, $S$ is the number of requests in the workload (the simulation time is proportional to the number of the requests) and $B$ is the beam size.
It runs reasonably fast for our medium-scale cluster when the number of requests is small. When the number of requests is very large, we propose another heuristic to accelerate: Instead of using the simulator to evaluate all (model, group) pairs at each iteration, we can run the simulator only once and place a model with the most unserved requests in an available group with the lowest utilization. This reduces the time complexity to $O((M+G)RS)$. We find this heuristic gives solutions with SLO attainment higher than 98\% of the SLO attainment get by the original algorithm in our benchmarks.

\begin{algorithm}[t!]
\caption{Enumeration-Based Group Partition and Model-Parallel Configuration Selection.}\label{alg:group-partition}
\begin{algorithmic}
\Require Model list $M$, cluster $C$, workload $W$.
\Ensure The placement $\mathit{best\_plm}.$
\State $\mathit{best\_plm} \leftarrow \emptyset$
\State $\mathcal{B} \leftarrow \text{get\_potential\_model\_buckets}(M)$
\For{$(B_1, B_2, \ldots, B_k) \in \mathcal{B}$}
  \State $\mathcal{H} \leftarrow \text{get\_potential\_device\_buckets}(C, B, k)$
  \For{$(H_1, H_2, \ldots, H_k) \in \mathcal{H}$}
     \State \textit{// Get the placement for each bucket individually.}
     \For{$i$ \textbf{from} $1$ \textbf{to} $k$}
        \State $\mathit{plm}_i^* \leftarrow \emptyset$
        \State $\mathcal{G} \leftarrow \text{get\_potential\_group\_partitions}(H_i)$
        \For{$G \in \mathcal{G}$}
            \State $\mathcal{P} \leftarrow \text{get\_potential\_parallel\_configs}(G)$
            \For{$P \in \mathcal{P}$}
                \State $\mathit{plm} \leftarrow \text{greedy\_selection}(B_i, G, P, W)$
                \If{$\mathit{plm}.\mathit{slo\_att}$ > $\mathit{plm}_i^*.\mathit{slo\_att}$}
                    \State $\mathit{plm}_i^* \leftarrow \mathit{plm}$
                \EndIf
            \EndFor
        \EndFor
    \EndFor
    \State $\mathit{plm}^* \leftarrow \text{concat}(\mathit{plm}_1^*,..., \mathit{plm}_k^*)$
    \If{$\mathit{plm}^*.\mathit{slo\_att}$ > $\mathit{best\_plm}.\mathit{slo\_att}$}
        \State $\mathit{best\_plm} \leftarrow \mathit{plm}^*$
    \EndIf
  \EndFor
\EndFor
\State \textbf{return} $\mathit{best\_plm}$
\end{algorithmic}
\end{algorithm}

\cref{alg:group-partition} enumerates different group partitions and model-parallel configurations and picks the best one via multiple calls to \cref{alg:greedy-placement}.
When designing \cref{alg:group-partition}, the first phenomenon we notice is that putting small and large models in the same group causes convoy effects, where the requests of small models have to wait for the requests of large models and miss the SLO.
Therefore, in \cref{alg:group-partition}, we first cluster models into model buckets. Each bucket contains a set of models with relatively similar sizes and every model is assigned to one and only one bucket. Specifically, the function \verb|get_potential_model_buckets| returns all the possible model bucket partitions that separate models whose latency difference is larger than a threshold into different disjoint buckets. We then enumerate all the potential ways to assign the devices to each bucket in \verb|get_potential_device_buckets|. 

Because different buckets include a disjoint set of models, we can then figure out the optimal placement for each bucket individually. For each bucket, we enumerate possible ways to partition the devices in the bucket into several groups in \verb|get_potential_group_partitions| and enumerate the potential parallel configurations for each group with the method in \verb|get_potential_parallel_configs|.
We then call \cref{alg:greedy-placement} with \verb|greedy_placement| to place models in the model bucket to the groups in the device bucket.
We send the whole workload $W$ to \cref{alg:greedy-placement}, which ignores the requests that hit the models outside of the current bucket.
Finally, a complete solution is got by concatenating the solutions for all buckets. The algorithm returns the best solution it finds during the enumerative search process. 

Enumerating all possible choices can be slow, so we use the following heuristics to prune the search space.
Intuitively, we want the different buckets to serve a similar number of requests per second.
Therefore, we eliminate the bucket configurations with high discrepancies in the estimated number of requests it can serve per second for each bucket.
Additionally, in \verb|get_potential_group_partitions| and \verb|get_potential_parallel_configs|, we assume all groups have the same size and the same parallel configurations except for the last group which is used when the number of devices is not divisible by the group size.

\subsection{Runtime Scheduling}
\label{subsec:runtime-schedulig}
We use a simple policy to dispatch and schedule the requests at runtime.
All requests are sent to a centralized controller. The controller dispatches each request to the group with the shortest queue length.
Each group manages a first-come-first-serve queue. When a group receives a request, it checks whether it can serve the request under SLO and rejects the request if it cannot. This is possible because the execution time of a DNN model is very predictable and can be got in advance by profiling~\cite{gujarati2020serving}.
In most of our experiments, we do not include advanced runtime policies such as batching~\cite{gujarati2020serving}, swapping, and preemption~\cite{han2022microsecond}. These techniques are complementary to model parallelism. Nevertheless, we discuss how they fit into our system.

\heading{Batching.} Batching multiple requests of the same model together can increase the GPU utilization and thus increase the throughput of a serving system. In our system, we do find batching is helpful, but the gain is limited. This is because we mainly target large models and a small batch size can already fully saturate the GPU, which is verified in \S\ref{subsec:batching-exp}. To isolate the benefits of model parallelism and make the results more explainable, we decide to disable any batching in this paper except for the experiments in \S\ref{subsec:batching-exp}.

\heading{Preemption.} The optimal scheduling decision often depends on future arrivals, and leveraging preemption can help correct previous suboptimal decisions. The first-come-first-serve policy may result in convoy effects when models with significantly different execution times are placed in the same group.
We anticipate a least-slack-time-first policy with preemption can alleviate the problems~\cite{davis1993scheduling}.

\heading{Swapping.} The loading overheads from the CPU or Disk to GPU memory are significant for large models, which is the target of this paper, so we do not implement swapping in \sys.
We assume all models are placed on the GPUs. This is often required due to tight SLOs and high rates, especially for large models. The placement of models in \sys can be updated in the periodic re-placement (e.g., every 24 hours).

\heading{Fault tolerance.} While the current design of \sys does not have fault tolerance as a focus, we acknowledge several potential new challenges for fault tolerance: With model parallelism, the failure of a single GPU could cause the entire group to malfunction. Additionally, the use of a centralized controller presents a single point of failure. 

\section{Implementation}
\label{sec:implementation}
We implement a real system and a simulator for \sys with about 4k lines of code in Python.
The real system is implemented on top of an existing model-parallel training system, Alpa~\cite{zheng2022alpa}.
We extend its auto-parallelization algorithms for inference settings to get the model-parallel strategies.
We then launch an Alpa runtime for each group and dispatch requests to these groups via a centralized controller.

The simulator is a continuous-time, discrete-event simulator~\cite{robinson2014simulation}.
The simulator maintains a global clock and simulates all requests and model executions on the cluster. Because the simulator only models discrete events, it is orders of magnitude faster than the real experiments. In our experiment, it takes less than 1 hour for a 24-hour trace. The fidelity of the simulator is very high because of the predictability of DNN model execution, which is verified in \S\ref{subsec:exp-setup}.

%% file: sec.eval.tex
\newcommand{\baseline}[0]{Clockwork++\xspace}

\section{Evaluation}
\label{sec:evaluation}
In this section, we evaluate \sys's serving ability under a variety of model and workload conditions.
The evaluation is conducted on a range of model sizes, including those that do and do not fit into a single GPU, and we show that \sys consistently outperforms strong baselines across all model sizes.
In addition, we evaluate the robustness of \sys against changing arrival patterns and do ablation studies of our proposed techniques.
Evaluation results show that \sys can greatly improve various performance metrics. Specifically, \sys can choose to save up to $2.3\times$ devices, handle $10\times$ higher rates, $6\times$ more burstiness, or $2.5\times$ more stringent SLO, while meeting the latency SLOs for over 99\% requests.

\subsection{Experiment Setup}
\label{subsec:exp-setup}
\topheading{Cluster testbed.}
We deploy \sys on a cluster with 8 nodes and 64 GPUs. Each node is an AWS EC2 p3.16xlarge instance with 8 NVIDIA Tesla V100 (16GB) GPUs.

\heading{Model setup.}
Since Transformer~\cite{vaswani2017attention} is the default backbone for large models, we choose two representative large Transformer model families: BERT~\cite{devlin2018bert} and GShard MoE~\cite{lepikhin2020gshard} for evaluation.\footnote{In this paper, we focus on non-autoregressive large models which perform inference with one forward pass, but note that the techniques proposed in this paper can be extended to auto-regressive models like GPT-3.}
In ML practice, the large model weights are usually pretrained and then finetuned into different versions for different tasks. Hence, for each model family, we select several most commonly used model sizes~\cite{brown2020language}, and then create multiple model instances at each size for experimentation. Also, we design some model sets to test the serving systems under different model conditions; details about model sizes, their inference latency on testbed GPUs, and the number of model instances in each model set are provided in Tab.~\ref{table:model_config}.

\begin{table}
\centering
\footnotesize
\begin{tabular}{lllllll}
\toprule
\textbf{Name} & \textbf{Size} & \textbf{Latency (ms)} & S1 & S2 & S3 & S4 \\
\midrule
BERT-1.3B & 2.4 GB & 151 & 32 & 0 & 10 & 0 \\
BERT-2.7B & 5.4 GB & 238 & 0 & 0 & 10 & 0 \\
BERT-6.7B & 13.4 GB & 395 & 0 & 32 & 10 & 0 \\
BERT-104B & 208 GB & 4600 & 0 & 0 & 0 & 4 \\
MoE-1.3B & 2.6 GB & 150 & 0 & 0 & 10 & 0 \\
MoE-2.4B & 4.8 GB & 171 & 0 & 0 & 10 & 0 \\
MoE-5.3B & 10.6 GB & 234 & 0 & 0 & 10 & 0 \\
\bottomrule
\end{tabular}
\caption{The first three columns list the sizes and inference latency of the models. The latency is measured for a single query with a sequence length of 2048 on a single GPU. BERT-104B's latency is reported using a minimal degree of inter-op parallelism. The latter columns list the number of instances for each model in different model sets named as S1-S4.
}
\vspace{-3mm}

\label{table:model_config}
\end{table}

\heading{Metrics.}
We use \emph{SLO attainment} as the major evaluation metric.
Under a specific SLO attainment goal (say, 99\%), we concern with another four measures: (1) the minimal number of devices the system needs, (2) the maximum average request rate, (3) the maximum traffic burstiness the system can support, and (4) the minimal SLO the system can handle. We are particularly interested in a SLO attainment of $99\%$ (indicated by vertical lines in all curve plots),  but will also vary each variable in (1) - (4) and observe how the SLO attainment changes.

\heading{Simulator fidelity.}
We want to study the system behavior under extensive models, workload, and resource settings, but some settings are just beyond the capacity of our testbed. Also, it is cost- and time-prohibitive to perform all experiments on the testbed for the days-long real traces. To mitigate the problem,  we use the simulator introduced in \S\ref{sec:implementation} for the majority of our experiments, noticing that DNN model execution~\cite{gujarati2020serving} has high predictability, even under parallel settings~\cite{zheng2022alpa,lepikhin2020gshard}.
We study the fidelity of the simulator in Tab.~\ref{table:simulator_fidelity}. Given two model placement algorithms, we compare the SLO attainment reported by the simulator and by real runs on our testbed under different \emph{SLO Scales}. The error is less than 2\% in all cases, verifying the accuracy of our simulator. Additionally, we conduct experiments on cluster testbed for results in \S\ref{sec:evaluation:large_models}.

\begin{table}
\footnotesize
\centering
\begin{tabular}{c|cc|cc}
\toprule
\multirow{2}{*}{\begin{tabular}[c]{@{}c@{}}SLO\\ Scale\end{tabular}} & \multicolumn{2}{c|}{Selective Replication} & \multicolumn{2}{c}{\sys} \\  \cline{2-3} \cline{4-5}
      & Real System & Simulator & Real System & Simulator \\ \midrule
0.5x  & 00.0\%      & 00.0\%    & 33.3\%      & 33.3\%    \\
1x    & 00.0\%      & 00.0\%    & 53.5\%      & 53.2\%    \\
1.5x  & 29.7\%      & 30.2\%    & 64.1\%      & 64.7\%    \\
2x    & 36.9\%      & 36.8\%    & 79.0\%      & 80.6\%    \\
3x    & 49.5\%      & 48.5\%    & 91.4\%      & 92.1\%    \\
4x    & 58.6\%      & 57.8\%    & 96.4\%      & 96.5\%    \\
5x    & 64.9\%      & 64.0\%    & 97.6\%      & 97.9\%    \\
10x   & 83.1\%      & 82.6\%    & 100.0\%      & 99.7\%    \\
\bottomrule
\end{tabular}
\caption{Comparison of the SLO attainment reported by the simulator and the real system under different SLO scales.}
\vspace{-4mm}
\label{table:simulator_fidelity}
\end{table}

\begin{figure*}[t]
\centering
\includegraphics[width=\linewidth]{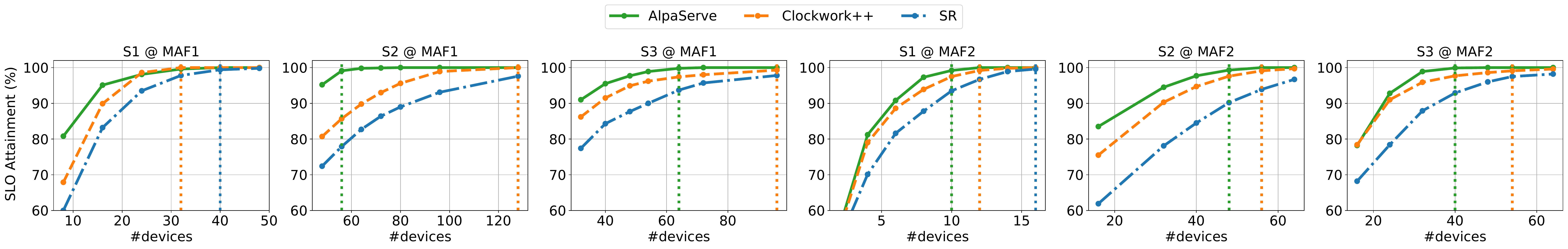}
\includegraphics[width=\linewidth]{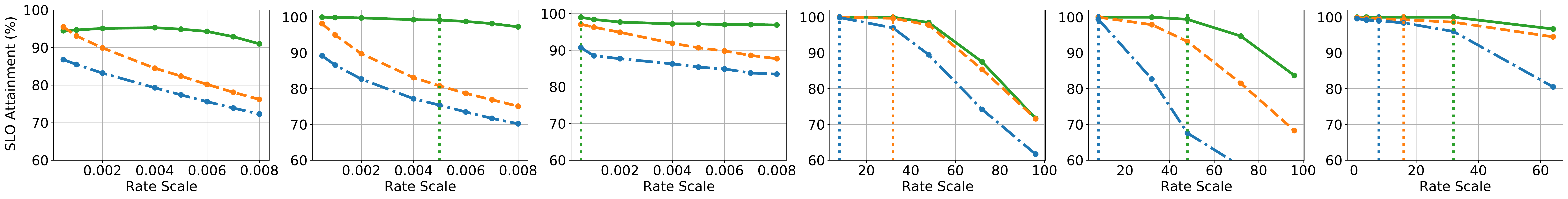}
\includegraphics[width=\linewidth]{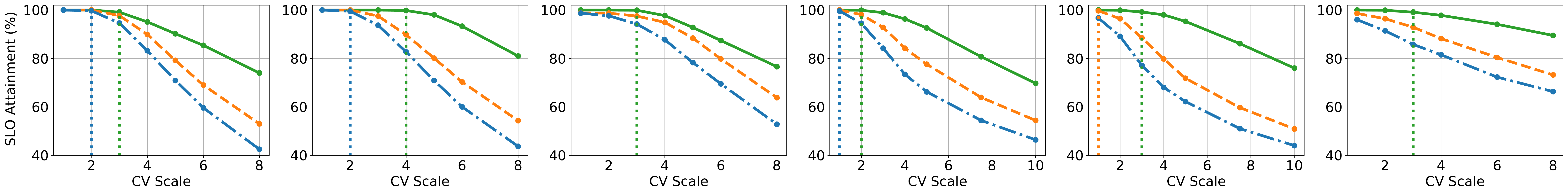}
\includegraphics[width=\linewidth]{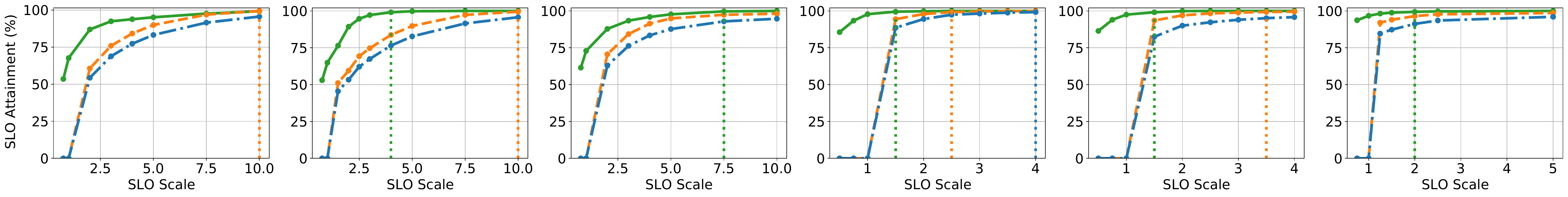}
\vspace{-5mm}
\caption{SLO attainment under various settings. In column S1@MAF1, we replay the MAF1 trace on the model set S1, and so on. In each row, we focus on one specific metric mentioned in \S\ref{sec:e2e-results} to see how its variation affects the performance of each serving system. If any, the dotted vertical line shows when the system can achieve 99\% SLO attainment.}
\vspace{-2mm}
\label{fig:realistic_workloads_results}
\end{figure*}

\subsection{End-to-end Results with Real Workloads}
\label{sec:e2e-results}

In this section, we compare \sys against baseline methods on publicly available real traces.

\heading{Workloads.} There does not exist an open-source production ML inference trace to the best of our knowledge. Therefore, we use the following two production traces as a proxy: Microsoft Azure function trace 2019 (MAF1)~\cite{shahrad2020serverless} and 2021 (MAF2)~\cite{zhang2021faster}. They were originally collected from Azure serverless function invocations in two weeks, and have been repurposed for ML serving research~\cite{bhattacharjee2019barista, ishakian2018serving}. 
The two traces exhibit distinct traffic patterns. In MAF1, each function receives steady and dense incoming requests with gradually changing rates; in MAF2, the traffic is very \emph{bursty} and is distributed across functions in a highly \emph{skewed} way -- some function receives orders of magnitude more requests than others.
Note that most previous works~\cite{gujarati2020serving} are evaluated on MAF1 only. 
Since there are more functions than models, following previous work~\cite{bhattacharjee2019barista,ishakian2018serving}, given a model set from Tab.~\ref{table:model_config}, we round-robin functions to models to generate traffic for each model. 

\heading{Setup.} SLO attainment depends on many factors. 
For each metric (1) - (4) mentioned in \S\ref{subsec:exp-setup}, we set a default value, e.g., the default SLO is set as tight as $5\times$ inference latency (SLO Scale=5). This forms a \emph{default setting}, given which,
we then vary one variable (while fixing others) at a time and observe how it affects the resulting SLO attainment. 
To change the two variables (3) and (4), which characterize traffic patterns, we follow Clockwork~\cite{gujarati2020serving} and Inferline~\cite{crankshaw2020inferline} and slice the original traces into time windows, and fit the arrivals in each time window with a Gamma Process parameterized by rate and coefficient of variance (CV). By scaling the rate and CV and resampling from the processes, we can control the rate and burstiness, respectively.

\heading{Baselines.} 
We compare \sys to two baseline methods: 
(1) \emph{Selective Replication (SR)}: use \sys's placement algorithm without model parallelism, which mimics the policy of a wide range of existing serving systems~\cite{crankshaw2017clipper, shen2019nexus};
(2) \emph{Clockwork++}: an improved version of the state-of-the-art model serving system Clockwork~\cite{gujarati2020serving}. The original Clockwork continuously swaps models into and out of GPUs. This helps for very small models (e.g., w/ several million parameters) but incurs significant swapping overheads on larger models. For fair comparisons, we implement \baseline in our simulator, which swaps models following Clockwork's replacement strategy at the boundary of every two windows\footnote{For MAF1, we follow Clockwork to set the window size as 60 seconds. For MAF2, we set it as 5.4K seconds.} of the trace using SR's algorithm, but \emph{assuming zero swapping overheads}. We believe it represents a hypothetical upper bound of Clockwork's performance.
Since all the baselines can only support models that can fit into one GPU memory,\footnote{In our cluster testbed, the per-GPU memory is 16GB, but the actual available space for model weights is around 13GB due to the need to store activations and other runtime context.} we use model set S1, S2 and S3 from Tab.~\ref{table:model_config} in this experiment.

\heading{SLO attainment vs. cluster size.}
\cref{fig:realistic_workloads_results}'s first row shows the SLO attainment with varying cluster sizes when serving a specific (model set, trace) pair. \sys outperforms the two baselines at all times and uses far fewer devices to achieve 99\% SLO attainment thanks to model parallelism. By splitting one model replica onto $N$ devices, \sys can achieve similar throughput as if $N$ replica were created for replication-only methods; but note \sys uses only one replica of memory.
Surprisingly, although we let \baseline adjust to the traffic dynamically with zero overhead, \sys still wins with a static placement; this is because model-parallel placement is by nature more robust to bursty traffic.

It is worth noting that replication-only methods can at most place 2 replicas of BERT-2.6B on a V100 (13GB memory budget), resulting in a substantial memory fraction, while model parallelism can avoid such memory fractions and enable more flexible placement of models.

\heading{SLO attainment vs. rate.}
\cref{fig:realistic_workloads_results}'s 2nd row varies the rate of the workloads. For a stable trace like MAF1, \sys can handle a much higher rate than baselines. While for a skewed and highly dynamic trace MAF2, whose traffic is dominated by a few models and changes rapidly, the replication-based methods have to allocate the majority of the GPUs to create many replicas for ``hot'' models to combat their bursty traffic; those GPUs, however, may go idle between bursts, even with frequent re-placement as in Clockwork++. In \sys, each model needs fewer replicas to handle its peak traffic. 

\heading{SLO attainment vs. CV.}
\cref{fig:realistic_workloads_results}'s 3rd row varies the CV of the workloads. The traffic becomes more bursty with a higher CV, which aggravates the queuing effect of the system and increases the possibility of SLO violation. The traditional solution to handle burstiness is by over-provision, wasting a lot of resources. \sys reveals a hidden opportunity to handle this by model parallelism. 

\heading{SLO attainment vs. SLO.}
\cref{fig:realistic_workloads_results}'s 4th row shows the effect of different SLO. Previous work~\cite{gujarati2020serving} which targets serving small models usually sets SLO to hundreds of milliseconds, even though the actual inference latency is less than 10 ms. Thanks to the intra-op parallelism, \sys can maintain good performance under similar SLO when serving large models, whose inference latency can be over 100 ms. When SLO is tight, even less than the model inference time, \sys favors intra-op parallelism to reduce the inference latency, which also reduces \sys's peak throughput due to the communication overhead but can make more requests to meet their SLO. When SLO becomes looser, \sys will automatically switch to use more inter-op parallelism to get higher throughput.

\subsection{Serving Very Large Models}
\label{sec:evaluation:large_models}
Today's large models may possess hundreds of billions of parameters~\cite{brown2020language,narayanan2021megatron,zhang2022opt}. To serve large models at this scale, the common practice in production is to choose the model parallelism strategy manually and use dedicated GPUs for each model~\cite{yu2022orca}. 
To show \sys has improved capability in serving very large models, we deploy model set S4 on our testbed, each requiring at least 16 GPUs to serve in terms of memory usage. As baselines, for each model, we enumerate all combinations of inter- and intra-op parallelisms on 16 GPUs. In contrast, \sys searches for the optimal GPU group allocation and model placement according to the arrival traffic and tries to achieve statistical multiplexing. 

\heading{Offered load.} In the default setting, the traffic is generated via a Gamma Process with an average rate of 8 requests/s and CV of 4. We then split the requests to each model following a power law distribution with an exponent of 0.5 to simulate the real-world skewness.\footnote{Uniform split yielded similar results.} Similar to \S\ref{sec:e2e-results}, we vary one of the rate, CV, or SLO in the default setting to see how each factor contributes to the resulting performance. It is worth noting that all results presented in this section are obtained via real execution on the testbed cluster.

\begin{figure}
    \centering
    \includegraphics[width=0.45\textwidth]{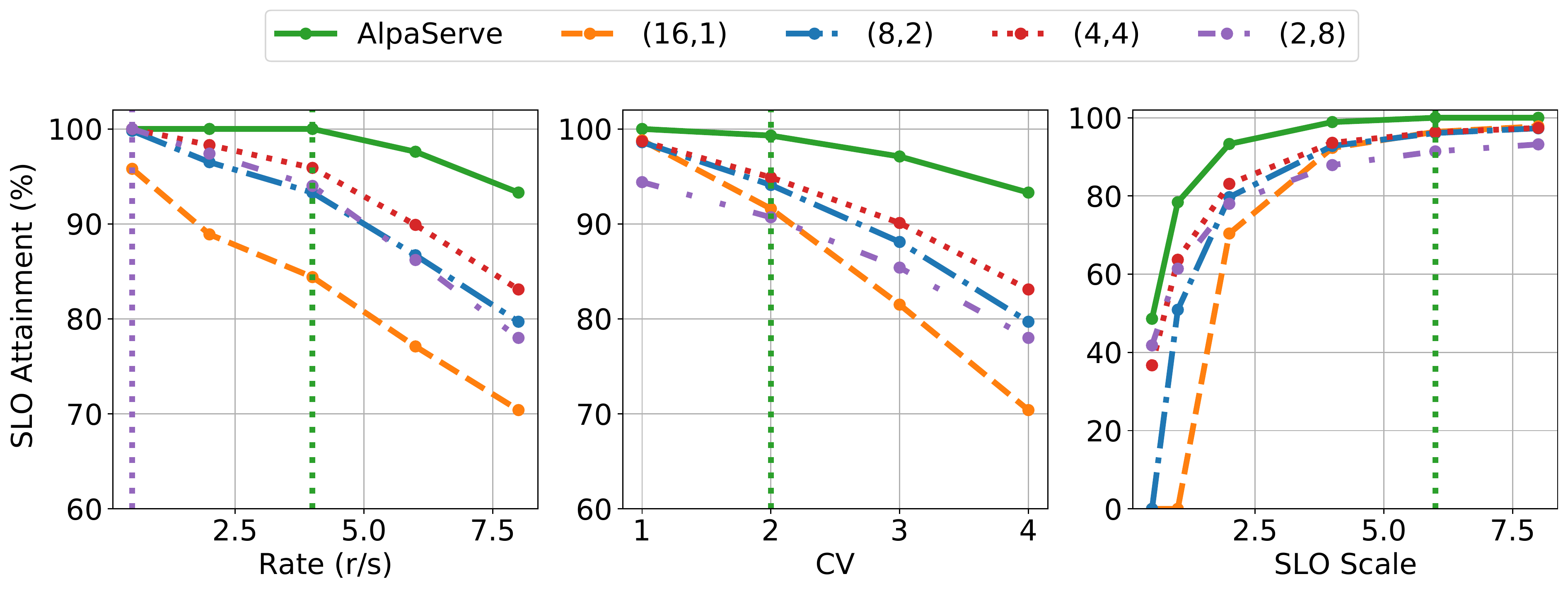}
    \vspace{-2mm}
    \caption{SLO attainment as we vary the rate, CV, and SLO scale. (8,2) means 8-way inter-op parallelism and in each pipeline stage using 2-way intra-op parallelism.}
    \vspace{-2mm}
    \label{fig:large_results}
\end{figure}

\heading{SLO attainment.} \cref{fig:large_results} shows the SLO attainment of each system under various settings. Although enumerating parallelism strategies and selecting the best can improve performance, it still remains a substantial gap compared to \sys. This means that the traditional way of using dedicated GPUs to serve large models is not ideal. We check the solution of \sys and find it slices the cluster evenly into two groups, each with the (4, 8) inter-/intra-op parallel configuration, and groups the models in a way that balances the requests between two groups. This further proves that our motivation in \S\ref{subsec:illustrative-example} still holds for extremely large models. By space-sharing the devices, \sys can exploit new opportunities for statistical multiplexing, which is advantageous for bursty workloads but largely under-explored by prior work.

\begin{figure*}
    \centering
    \includegraphics[width=.9\textwidth]{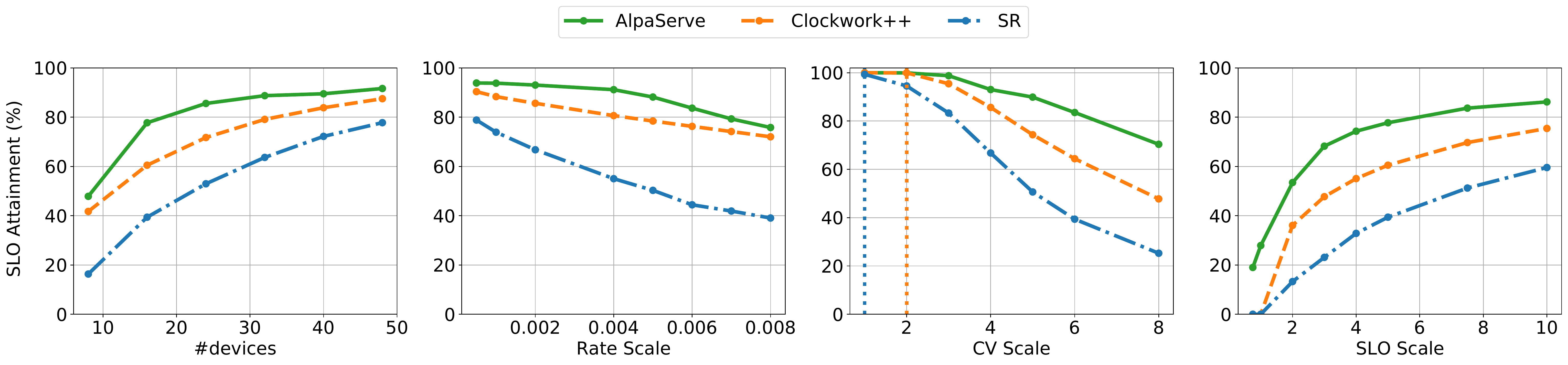}
    \vspace{-2mm}
    \caption{The actual arrival traffic for \sys and SR is different from what their algorithms are assumed, while \baseline runs directly on the actual traffic.}
    \vspace{-2mm}
    \label{fig:robustness_results}
\end{figure*}

\subsection{Robustness to Changing Traffic Patterns}
Until now, \sys's good performance is based on the assumption we make in its placement algorithm that we know the arrival process in advance. In practice, the arrival process can be approximated using historical traces but the unavoidable real-world variance may make the prediction inaccurate. In this experiment, we study how \sys performs if the traffic patterns change.

We reuse the same setting for S2@MAF1 in \S\ref{sec:e2e-results}, but this time for \sys and SR, we randomly slice two one-hour traces from MAF1, one is what their algorithms are assumed, while the other one is used as the actual arrival process. While for \baseline, we still run its algorithm directly on the actual arrival process to respect its online nature. Similarly, we vary different factors and compute the SLO attainment for each system. We repeat the experiments three times and show the average results in Fig.~\ref{fig:robustness_results}.

Unsurprisingly, SR's performance drops significantly when traffic changes. By contrast, \sys maintains good performance and still outperforms \baseline, an online adjustment algorithm, using a static placement generated from substantially different traffic patterns. This confirms that, in face of highly-dynamic traffic patterns, statistical multiplexing with model parallelism is a simple and better alternative than existing replication- or replacement-based algorithms.

\begin{figure}
    \centering
    \includegraphics[width=.46\textwidth]{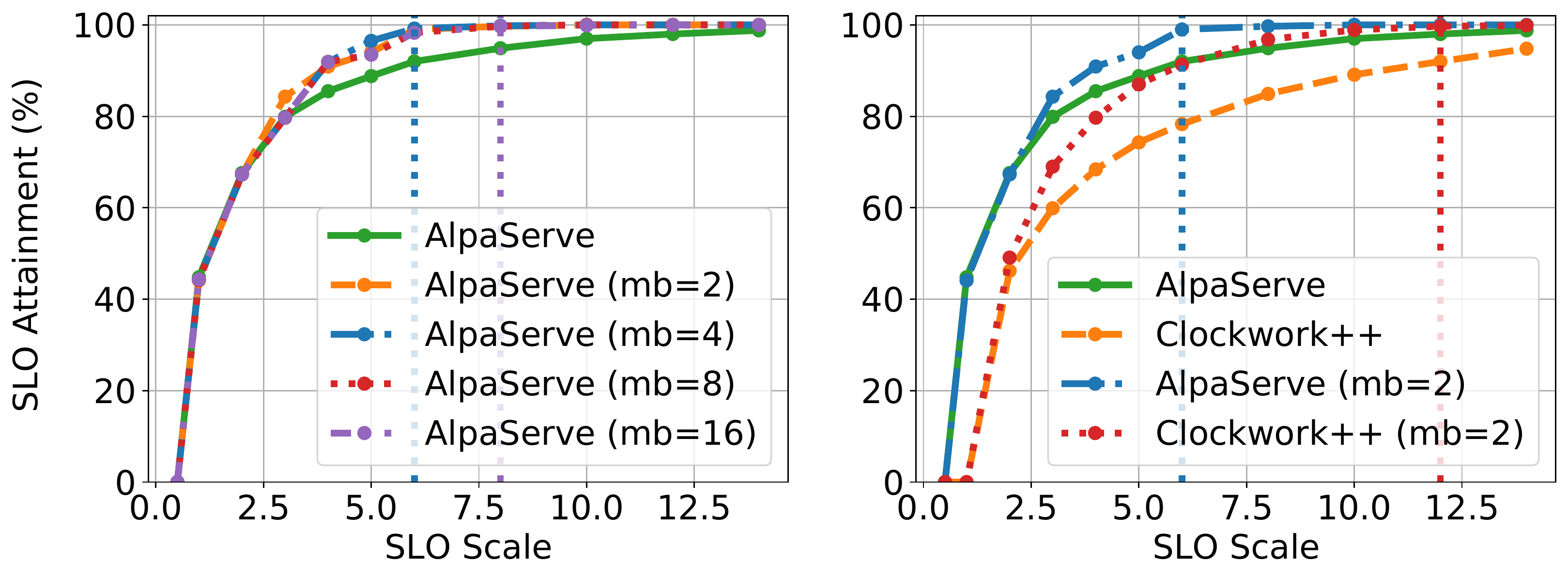}
    \caption{SLO Attainment when batching is enabled. mb=2 means the maximum batch size is 2.}
    \vspace{-2mm}
    \label{fig:batching_res}
\end{figure}

\subsection{Benefits of Dynamic Batching}
\label{subsec:batching-exp}
Batching is a common optimization in other serving systems~\cite{gujarati2020serving, nvidiatriton, olston2017tensorflow} and the choice of batch size is critical to the performance because it can increase GPU utilization and thus increase the system throughput. However, in large model scenarios, the benefit of batching is limited mainly due to two reasons. First, for large models, a small batch size will saturate the GPU, which means there is little gain to batching more requests. Second, the execution latency grows linearly with the batch size~\cite{shen2019nexus}, so when the SLO is tight (say SLO Scale is less than 2), batching is simply not a choice.

To isolate the benefits of model parallelism and make the results more explainable, we decide to disable any batching in other experiments but prove that the batching strategy is purely orthogonal to the scope of this paper in this subsection. To prove this, we implement a standard batching algorithm in \sys and evaluate its performance.

\heading{Batching strategy.} When a request arrives, it will get executed immediately if any device group is available. Otherwise, it will be put into a per-model requests queue for batching. When a device group becomes idle, it will choose a model which has a replica on it and batch as many requests as possible from the requests queue of the model while satisfying the SLO requirements.

\heading{Setup.} As the model size increases, the potential benefit of batching decreases. Therefore, we choose to evaluate model set S1. 
We generate a synthetic Gamma Process traffic with an average rate of 4 requests/s and a CV of 4 for each model.

\heading{SLO attainment.} \cref{fig:batching_res} (left) shows the SLO attainment achieved by \sys with different maximum batch size settings under various SLO scales. When the SLO requirement is tight, any batching will violate the SLO so there is no gain with batching enabled. Also, although we choose to serve the smallest model in Tab.~\ref{table:model_config}, a small batch size like 2 combined with a long sequence length of 2048 already saturates the GPU, so a larger maximum batch size brings no performance improvement. \cref{fig:batching_res} (right) compares the improvement for \sys and \baseline with our batching algorithm enabled.\footnote{SR is left out to make the figure clearer as it is worse than \baseline.} When the SLO requirement becomes loose, both \sys and \baseline have better SLO attainment to some extent. Since \sys's performance is good even without batching and batched requests with different batch sizes will incur stage imbalance and pipeline bubble in inter-op parallel, the absolute improvement of \baseline is slightly better.

\subsection{Ablation Study}
\label{sec:ablation}
In this section, we study the effectiveness of our proposed auto-parallelization (\S\ref{subsec:auto-parallel}) and placement algorithms (\S\ref{subsec:placement-alg}).

\begin{figure}
\centering
\vspace{-2mm}
     \begin{subfigure}[b]{0.23\textwidth}
     \centering
     \includegraphics[width=\textwidth]{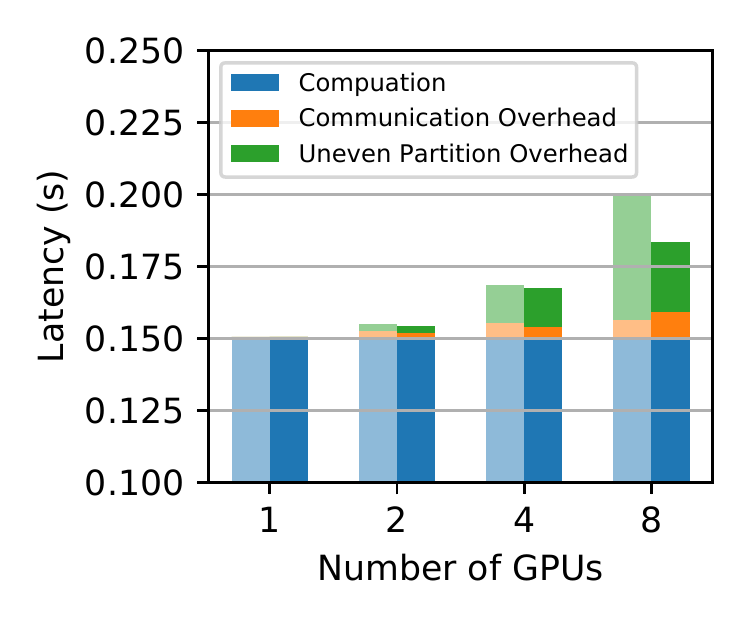}
         \vspace{-6mm}\caption{Transformer 1.3B.}
    \end{subfigure}
     \begin{subfigure}[b]{0.23\textwidth}
     \centering
     \includegraphics[width=\textwidth]{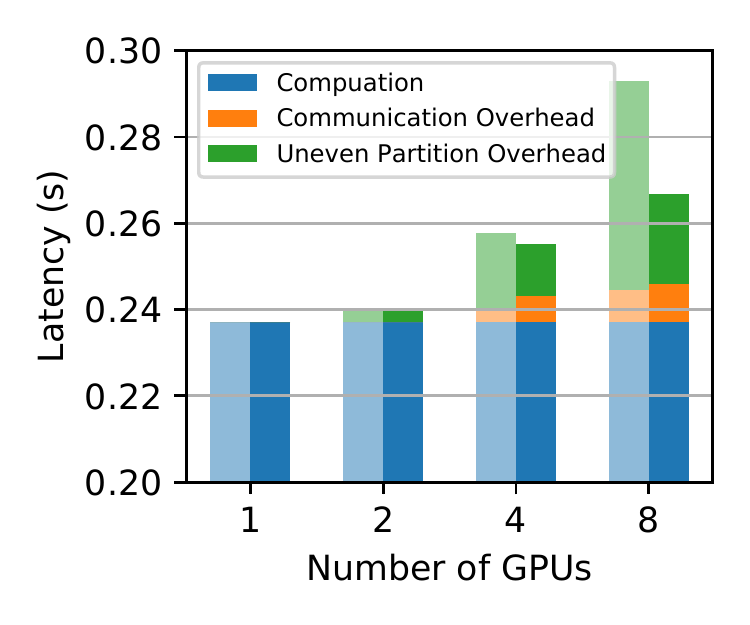}
         \vspace{-6mm}\caption{Transformer 2.6B.}
    \end{subfigure}
\vspace{-2mm} 
\caption{Comparison of the model parallel overhead between manual partition (lighter color) and the partition found by the automatic algorithm (darker color).} \vspace{-2mm}
\label{fig:ablation_auto_parallel}
\end{figure}

\heading{Benefits of auto-parallelization.}
We show that the auto-parallelization ability allows \sys to not only generalize to arbitrary model architectures but even also reduce parallelism overheads -- hence improved serving performance (see \S\ref{subsec:model-parallel-overhead} for more discussion).
To see that, typical manual model-parallel parallelization strategies offered in \emph{de facto} systems~\cite{aminabadi2022deepspeed, narayanan2021megatron, nvidiaft} is to assign an equal number of (transformer) layers to each pipeline stage. These strategies often fail to create balanced workloads across distributed GPUs because contemporary large models have heterogeneous layers, such as embedding operations. The extensions introduced in \S\ref{subsec:auto-parallel} automatically partition the models at the computational graph level and generate nearly-balanced stages. Empirically, as shown in Fig.~\ref{fig:ablation_auto_parallel}, for 8 pipeline stages, auto-parallelization reduces the total overhead by 32.9\% and 46.7\% for Transformer 1.3B and 2.6B respectively, which is necessary for achieving good serving performance when model parallelism is used for serving.

\begin{figure}
    \centering
    \includegraphics[width=0.45\textwidth]{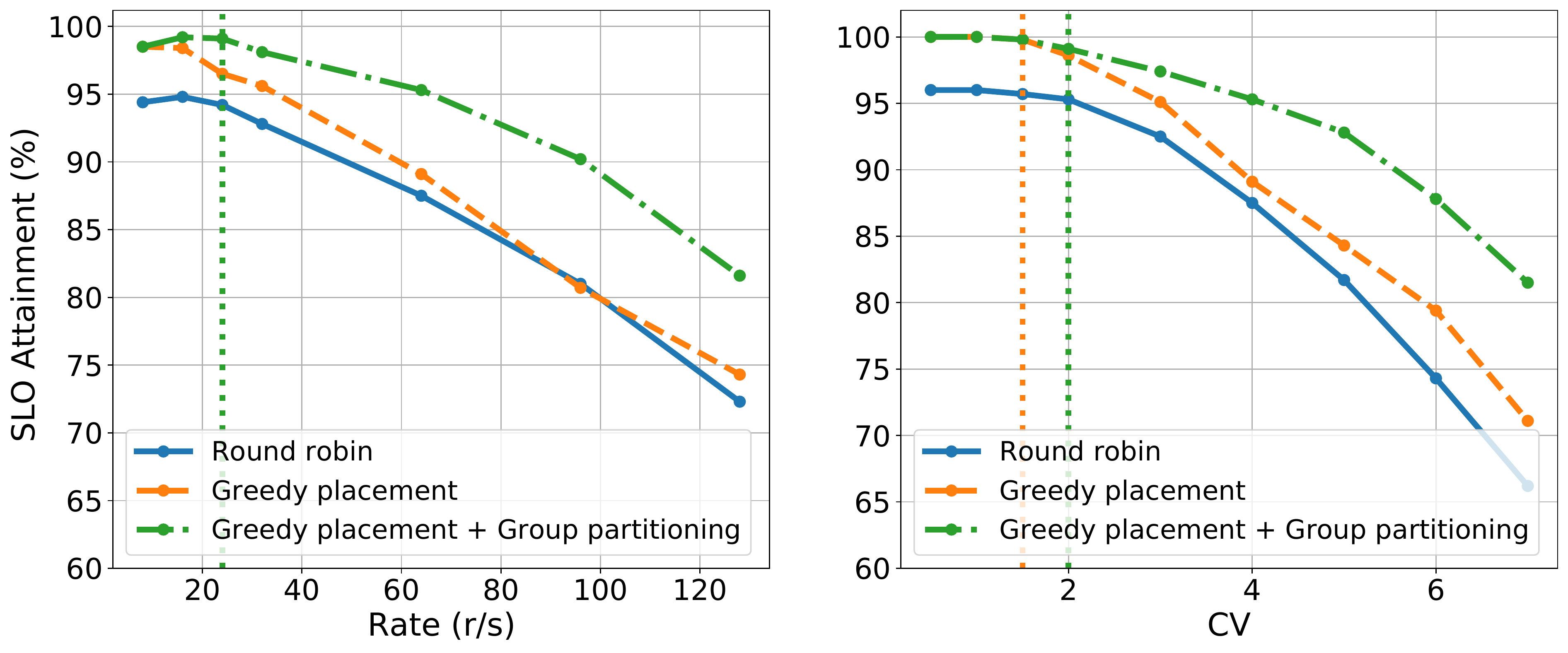}
    \vspace{-2mm}
    \caption{Ablation study of placement algorithms.}
    \vspace{-5mm}
    \label{fig:ablation-mixed}
\end{figure}

\heading{Effectiveness of the placement algorithm.}
We now test the effectiveness of our placement algorithm on a synthetic workload. We serve the most challenging model set S3 (Tab.~\ref{table:model_config}) on our testbed. The rate distribution of the models follows a power law distribution. The arrival pattern of each model is a Gamma process. Three variants of the placement algorithms are evaluated. \emph{Round robin} means placing models in a round-robin fashion and using 4-stage pipelines for all groups. \emph{Greedy placement} uses our greedy placement and 4-stage pipeline for all groups. \emph{Greedy placement + Group partitioning} performs greedy placement plus group partitioning search. As shown in Fig.~\ref{fig:ablation-mixed}, both placement and group partitioning are necessary to achieve good SLO attainment.
In the left subfigure, the group partitioning increases the rate by 1.5$\times$ compared to greedy placement without group partitioning over 99\% SLO attainment, while round robin can never reach 99\% SLO attainment.
In the right subfigure, the group partitioning increases the traffic burstiness that can be handled to meet 99\% SLO attainment by $1.3\times$.

%% file: sec.related_work.tex
\section{Related Work}

\topheading{Model serving systems.}
There has been a proliferation of model serving systems recently.
These range from general-purpose production-grade systems like TensorFlow Serving~\cite{olston2017tensorflow} and NVIDIA Triton~\cite{nvidiatriton}, which are widely used but do not provide any support for automatic placement or latency constraints.
They also include systems that are optimized for single-model serving~\cite{yu2022orca} or serving of specific classes of models (e.g., transformers)~\cite{yu2022orca,fang2021turbotransformers,zhou2022pets}.
\sys targets a broader set of models and features than these systems.

For SLO-aware, distributed serving, most serving systems ignore placement-level interactions between models.
Clockwork~\cite{gujarati2020serving}, for instance, primarily focuses on predictability; when scheduling, it greedily loads and executes models on available GPUs. Shepherd~\cite{zhang2023shepherd} utilizes preemption to correct sub-optimal scheduling decisions. For large models, loading model weights and preemption can easily overwhelm practical SLOs.
Other systems like Clipper~\cite{crankshaw2017clipper}, Infaas~\cite{infaas}, and DVABatch~\cite{cui2022dvabatch} also do not reason about the latencies of co-located models.

Nexus~\cite{shen2019nexus} is very related to our work in that it examines the placement of models;
however, Nexus is an example of a system that takes the traditional replication approach described in \S\ref{sec:motivation} and, thus, misses a broad class of potential parallelization strategies that we explore in this paper.

\heading{Inference optimizations for large models.}
\sys is complementary to another large body of work on optimizations for inference over large models.
These include techniques like quantization~\cite{dettmers2022llm}, distillation~\cite{sanh2019distilbert}, offloading~\cite{aminabadi2022deepspeed}, better operator parallelism~\cite{pope2022efficiently}, and CUDA kernel optimization~\cite{dao2022flashattention,ivanov2021data}.
Some of these optimizations are intended to stem the tide of increasing model sizes; however, all of these gains are partial--- the challenge of serving large models has continued to escalate rapidly despite these efforts.

\heading{Model parallelism for training.}
\sys is largely orthogonal to the large body of work on model parallelism in training~\cite{zheng2022alpa,narayanan2021megatron,huang2019gpipe,rajbhandari2020zero,li2021terapipe}.
As described in \S\ref{sec:motivation}, serving presents a unique set of constraints and opportunities not found in training workloads.
Where these systems do intersect with \sys, however, is in their methods for implementing model parallelism along various dimensions. In particular, \sys builds on some of the parallelization techniques introduced in~\cite{zheng2022alpa}.

\heading{Resource allocation and multiplexing.}
The problem of how to multiplex limited resources to the incoming requests is one of the oldest topics in computer science and has been studied in different application domains~\cite{rashmi2016ec,meng2010efficient,baruah2022parallelism}. Recent work on DL scheduling uses swapping~\cite{bai2020pipeswitch}, preemption~\cite{preemption22}, interleaving~\cite{zhao2022multi}, and space-sharing~\cite{TGS} to realize fine-grained resource sharing.
Rather, the contribution of this paper is a deep empirical analysis of the applications of these ideas to an emerging space: the serving of multiple large models.

%% file: sec.conclusion.tex
\section{Conclusion and Future Work}
In this paper, we presented \sys, a system for prediction servings of multiple large deep-learning models. The key innovation of \sys is integrating model parallelism into multi-model serving.
Because of the inherent overheads of model parallelism, such parallelism is traditionally applied conservatively---reserved for cases where models simply do not fit within a single GPU or execute within the required SLO.
\sys demonstrates that model parallelism is useful for many other scenarios, quantifies the tradeoffs, and presents techniques to automatically navigate that tradeoff space.

In the future, we will extend \sys to more complicated scenarios, including serving multiple parameter-efficient adapted models (e.g., LoRA \cite{hu2021lora}), models with dependencies, and autoregressive models \cite{brown2020language}.

\section{Acknowledgement}
We thank the OSDI reviewers and our shepherd, Heming Cui, for their valuable feedback. This work
is in part supported by NSF CISE Expeditions Award CCF1730628, NSFC under the grant number 62172008, and gifts from Astronomer, Google, IBM, Intel, Lacework, Microsoft, Nexla, Samsung SDS, Uber, and VMware.
Yinmin Zhong and Xin Jin are also with the Key Laboratory of High Confidence Software Technologies (Peking University), Ministry of Education.